\documentclass[sigconf,nonacm]{acmart}

\acmSubmissionID{1234}

\AtBeginDocument{%
  \providecommand\BibTeX{{%
    \normalfont B\kern-0.5em{\scshape i\kern-0.25em b}\kern-0.8em\TeX}}}
    
\usepackage{graphicx}
\usepackage{amsmath,amsfonts}
\usepackage{svg}
\usepackage{hyperref}
\usepackage{multirow}
\usepackage{tabularx}
\usepackage{colortbl}
\usepackage{xcolor}
\usepackage{booktabs} % for better table rules
\usepackage{textcomp}
\usepackage{booktabs}
\usepackage{rotating} % for rotated text
\usepackage{array} % for fixed width columns
\usepackage{adjustbox}
\usepackage{siunitx}
\usepackage{tikz}
\usepackage{subcaption}
\usepackage{threeparttable}
\usepackage{stfloats}

\setcopyright{acmlicensed}
\copyrightyear{2024}
\acmYear{2024}
\acmDOI{XXXXXXX.XXXXXXX}

\acmConference[CIKM 2024]{Make sure to enter the correct
  conference title from your rights confirmation email}{June 03--05,
  2018}{Woodstock, NY}
\acmISBN{978-1-4503-XXXX-X/18/06}

\DeclareMathOperator{\agg}{AGGREGATE}
\DeclareMathOperator{\merge}{MERGE}
\DeclareMathOperator{\pool}{POOL}
\DeclareMathOperator{\comb}{COMBINE}
\DeclareMathOperator{\upd}{UPDATE}
\DeclareMathOperator{\gnn}{GNN}

\newcommand{\confone}{\textnormal{SE2P-C1}}
\newcommand{\conftwo}{\textnormal{SE2P-C2}}
\newcommand{\confthree}{\textnormal{SE2P-C3}}
\newcommand{\conffour}{\textnormal{SE2P-C4}}

\newcommand{\mutag}{\text{\small MUTAG}}
\newcommand{\proteins}{\text{\small PROTEINS}}
\newcommand{\ptc}{\text{\small PTC-MR}}
\newcommand{\imdbb}{\text{\small IMDB-B}}
\newcommand{\imdbm}{\text{\small IMDB-M}}
\newcommand{\collab}{\text{\small COLLAB}}
\newcommand{\ogbmolhiv}{\text{\small OGBG-MOLHIV}}
\newcommand{\ogbmoltox}{\text{\small OGBG-MOLTOX}}

\definecolor{customyellow}{rgb}{1, 0.65, 0}
\definecolor{customgreen}{rgb}{0, 0.65, 0}
\definecolor{customred}{rgb}{0.9, 0, 0}

\newcommand{\greencircle}{\protect\tikz \fill [customgreen] (0,0) circle (0.6ex);}
\newcommand{\redcircle}{\protect\tikz \fill [customred] (0,0) circle (0.6ex);}
\newcommand{\yellowcircle}{\protect\tikz \fill [customyellow] (0,0) circle (0.6ex);}

\begin{document}
\title{Scalable Expressiveness through Preprocessed Graph Perturbations}

\author{Danial Saber}
\email{danial.saber@ontariotechu.ca}
\affiliation{%
  \institution{Ontario Tech University}
  \city{Oshawa}
  \state{Ontario}
  \country{Canada}
}

\author{Amirali Salehi-Abari}
\email{abari@ontariotechu.ca}
\affiliation{%
  \institution{Ontario Tech University}
  \city{Oshawa}
  \state{Ontario}
  \country{Canada}
}

\renewcommand{\shortauthors}{Saber and Salehi-Abari}

\begin{abstract}
Graph Neural Networks (GNNs) have emerged as the predominant method for analyzing graph-structured data. However, canonical GNNs have limited expressive power and generalization capability, thus triggering the development of more expressive yet computationally intensive methods. One such approach is to create a series of perturbed versions of input graphs and then repeatedly conduct multiple message-passing operations on all variations during training. Despite their expressive power, this approach does not scale well on larger graphs. To address this scalability issue, we introduce \textit{Scalable Expressiveness through Preprocessed Graph Perturbation (SE2P)}. This model offers a flexible, configurable balance between scalability and generalizability with four distinct configuration classes. At one extreme, the configuration prioritizes scalability through minimal learnable feature extraction and extensive preprocessing; at the other extreme, it enhances generalizability with more learnable feature extractions, though this increases scalability costs. We conduct extensive experiments on real-world datasets to evaluate the generalizability and scalability of SE2P variants compared to various state-of-the-art benchmarks. Our results indicate that, depending on the chosen SE2P configuration, the model can enhance generalizability compared to benchmarks while achieving significant speed improvements of up to 8-fold. 
\end{abstract}


\settopmatter{printfolios=true}
\maketitle

\section{Introduction}
Graph Neural Networks (GNNs) possess exceptional predictive capabilities on relational data (e.g., social networks, protein-protein interaction networks, etc.). 
Their applicability spans various domains, such as recommender systems \cite{wu2022graph}, protein structure modeling \cite{gao2020deep}, educational systems \cite{namanloo2022improving}, and knowledge graph completion \cite{arora2020survey}. However, relational data's complexity, scale, and dynamic nature pose substantial challenges to GNNs, emphasizing the importance of improving their generalization and computational efficiency.

Message-passing GNNs (MPNNs), a popular class of GNNs, facilitate the exchange of messages between nodes to integrate their local structural and feature information within a graph. %This communication enables nodes to learn representations that capture both their local structure in the graph and their associated feature information. 
Stacking multiple layers of message-passing allows the messages to be transmitted across longer distances in the graph, thus learning more global structural information. However, this approach increases computational complexity, particularly for larger graphs, and its generalization capabilities are limited by the $1$-dimensional Weisfeiler-Lehman ($1$-WL) graph-isomorphism test \cite{leman1968reduction}.

Several approaches have been proposed to enhance the computational efficiency of message-passing GNNs. \textit{Simplified graph neural networks (SGNNs)} simplifies graph neural networks by removing their non-linearities between intermediate GNN layers, thus facilitating the preprocessing of message-passing for faster training \cite{wu2019simplifying,sign_icml_grl2020,louis2023simplifying,zhang2022nafs, feng2020graph}. Another acceleration approach involves sampling methods, where node neighborhoods are sampled during preprocessing \cite{chiang2019cluster,zeng2019graphsaint,shi2023lmc, louis2022sampling} or message-passing \cite{hamilton2017inductive,chen2018fastgcn,huang2018adaptive,zou2019layer,ying2018graph,chen2018stochastic} to reduce the number of messages during training or inference. The $1$-WL expressivity constraint also extends to these scalability approaches for message-passing GNNs.

To go beyond $1$-WL expressive power, many attempts are made. \textit{Higher-order GNNs}, similar to the hierarchy in $k$-WL tests \cite{li2019graph,maron2019provably}, expand message-passing from individual nodes to node tuples, yielding richer representations. As an alternative approach, \textit{Feature-augmented GNNs} integrate additional features like structural encodings \cite{bouritsas2022improving,barcelo2021graph}, geodesic distances \cite{li2020distance,zhang2021labeling,velingker2022affinity}, resistance distances \cite{velingker2022affinity}, and positional encodings \cite{dwivedi2022graph} for nodes or edges.
% Other approaches opt for enhanced standard MPNNs by incorporating structural encodings for nodes and edges, such as counting cycles or cliques \cite{chen2020can}.
Other approaches known as \textit{Subgraph GNNs} focus on extracting multiple subgraphs for each node, using techniques such as node/edge labeling \cite{you2021identity,papp2022theoretical,huang2022boosting,jacob2023stochastic} or node/edge deletion \cite{papp2021dropgnn,huang2022boosting,bevilacqua2021equivariant,rong2020dropedge}. These techniques generate a variety of subgraph perturbations, enriching the graph's representational diversity. 

Despite their improved expressiveness, these approaches are less computationally efficient than the conventional GNNs. Higher-order GNNs require (at least) cubic computational complexity for message passing  \cite{zhang2021nested,zhang2022knowledge}. Feature-augmented GNNs can struggle with scalability, as computing augmented features (e.g., structural encodings, resistance distance, shortest path distance, etc.) is computationally expensive for large graphs. Subgraph GNNs typically involve extracting or creating multiple large subgraphs from a larger graph and processing these with GNNs. Due to potential overlap among these subgraphs, their cumulative size can significantly expand, sometimes to hundreds of times the size of the original graph, rendering them impractical for large-scale graphs. For instance, DropGNN \cite{papp2021dropgnn} generates multiple perturbations of the graph through random node removal and applies GNN on the perturbed graphs during training to enhance expressivity. However, scalability becomes a critical concern as the number of perturbations grows, confining the method's applicability to smaller datasets. 
\vskip 1mm
\noindent \textbf{Our approach.} In our pursuit of formulating a flexible, scalable, and expressive model, we introduce \textit{Scalable Expressiveness through Preprocessed Graph Perturbation (SE2P)}. Our approach offers four configuration classes, each offering a unique balance between scalability and generalizability. SE2P first creates multiple perturbations of the input graph through a perturbation policy (e.g., random node removal) and then diffuses nodal features through each perturbed graph. Unlike conventional message-passing GNNs, but similar to SGNNs \cite{wu2019simplifying, sign_icml_grl2020,louis2023simplifying, zhang2022nafs, feng2020graph}, the feature diffusion occurs once during preprocessing. Our SE2P differentiates from SGNNs by leveraging the expressive power offered by multiple perturbed graphs \cite{papp2021dropgnn,bevilacqua2021equivariant}. The diffusions from each perturbation are \textit{combined}, and nodal representations of all perturbations are \textit{merged} to construct the final nodal profiles. Then, these profiles are subjected to a \textit{pooling} operation for graph classification tasks to produce a vector representation of the entire input graph. A critical computational strength of our framework is its flexibility, allowing for the selection of aggregation functions (learnable or non-learnable), thus enabling scalable or expressive variations of many models. Depending on the selected configuration, our empirical results demonstrate significant speedup (up to $8\times$) compared to baselines while also improving the generalizability of expressive models such as DropGNN. Our experiments confirm that SE2P surpasses the generalization limitation of SGNNs (e.g., SGCNs, and SIGN) while offering comparable scalability. SE2P demonstrates flexibility in balancing expressiveness and scalability: The SE2P's instance $\conftwo$ achieves $3$-$6\times$ speed up with comparable generalizability to baselines, while another instance $\confthree$ maintains baseline computational requirements and enhanced generalizability by up to $1.5\%$. 

\section{Related Work}
We review existing methods that enhance the scalability (i.e., computational efficiency) or expressiveness of Graph Neural Networks. We have identified key subdomains within these areas.
\vskip 0.5mm
\noindent\textbf{Simplification Methods (Scalability).} To reduce the computational burden associated with multiple message-passing layers, one strategy \cite{wu2019simplifying, sign_icml_grl2020, louis2023simplifying, feng2020graph, zhang2022nafs} is to simplify GNN architectures by removing intermediate non-linearities, thus enabling feature propagation preprocessing across the graph. \textit{Simplified Graph Convolutional Networks (SGCN)} \cite{wu2019simplifying} introduces a diffusion matrix by removing non-linearities from multi-layer Graph Convolutional Networks \cite{kipf2016semi}. SIGN \cite{sign_icml_grl2020}, GRAND \cite{feng2020graph}, and NAFS \cite{zhang2022nafs} extend SGCN to multiple diffusion matrices instead of a single one. Similarly, S3GRL \cite{louis2023simplifying} extends this technique to enrich the subgraph representations. However, many diffusion approaches (e.g., SGCN, SIGN, etc.) exhibit limited expressivity. Our SE2P framework leverages this feature diffusion approach while enhancing these models with graph perturbations to overcome their expressivity limitations. %\textit{Graph Random Neural Networks (GRAND)} leverages the same approach of combining a set of diffusion matrices, with the distinction of normalizing the feature information with the propagation step (i.e., power of the diffusion matrix). Our model's feature diffusion approach aligns with this strategy to enhance scalability but improves their limited expressiveness power. 
\vskip 0.5mm
\noindent\textbf{Sampling Methods (Scalability).}
The sampling techniques are deployed to control the exponential growth of receptive fields across GNN layers for better scalability. \textit{Node-based} sampling methods \cite{hamilton2017inductive,ying2018graph,chen2018stochastic} samples fixed-size nodes' neighborhoods for message-passing within mini-batches to reduce the nodes involved in the message-passing process. In contrast, \textit{layer-based} sampling methods \cite{chen2018fastgcn,huang2018adaptive,zou2019layer} targets the architectural level of GNNs by selectively sampling nodes for each layer. \textit{Subgraph-based} sampling \cite{chiang2019cluster,zeng2019graphsaint,shi2023lmc,louis2022sampling,jacob2023stochastic} focuses on sampling subgraphs during preprocessing to handle large graphs on limited memory, and integrates these smaller, representative subgraphs into minibatches during training to update embeddings. Despite scalability, these sampling methods are limited to the $1$-WL isomorphism test, necessitating enhancements in GNN expressivity.

\vskip 0.5mm
\noindent\textbf{Higher-order GNNs (Expressivity).} Inspired by the higher-order WL tests \cite{morris2017glocalized}, higher-order GNNs pass messages between node tuples instead of individual nodes \cite{morris2019weisfeiler,azizian2020expressive}. To offset their high computational demands, localized and sparse higher-order aggregation methods have been developed \cite{morris2020weisfeiler}, with the cost of reduced expressiveness \cite{morris2022speqnets}.  %Given the significant computational demands of higher-order GNNs, some attempts have focused on enhancing their computational efficiency, such as introducing more localized and sparse higher-order aggregation methods \cite{morris2020weisfeiler}. While this strategy enhances computational efficiency, it does come at the cost of reducing a certain level of expressiveness \cite{morris2022speqnets}.

%\vskip 0.5mm
%\noindent\textbf{Substructure-based GNNs (Expressivity).} Identifying specific substructures within a graph (e.g., cycles with specific lengths) is crucial for various predictive tasks, especially in computational chemistry \cite{koyuturk2004efficient}, biology \cite{irwin2012zinc}, and social network analysis \cite{jiang2010finding}. Canonical message-passing GNNs fail to correctly detect or enumerate prevalent substructures such as cycles, cliques, and paths \cite{chen2020can}. Therefore, GNNs are extended to recognize or count these substructures within a given graph \cite{bouritsas2022improving,%Horn22a,
%barcelo2021graph}. Graph substructure network (GSN) \cite{bouritsas2022improving} incorporates substructure counting into node features through a preprocessing step. However, this preprocessing step has an intractable worst-case complexity, and deciding on substructure features often requires domain-specific knowledge \cite{bevilacqua2021equivariant}. Similarly, \cite{barcelo2021graph} extends this approach by using homomorphism counts to incorporate various substructure counts (such as induced subgraphs, cliques, and cycles) into the feature representations. TOGL \cite{Horn22a} further developed this idea by designing a topology-aware aggregation layer that considers substructures and can be integrated into the GNN architecture.
\vskip 0.5mm
\noindent\textbf{Subgraph GNNs or Perturbed GNNs (Expressivity).} An emerging body of research has focused on generating a bag of subgraphs (or perturbations) from an input graph and then applying GNNs on all subgraphs to get more expressive nodal representations. The primary distinction between these approaches is how subgraphs are generated or sampled. Some common techniques include node dropping \cite{papp2021dropgnn,cotta2021reconstruction,feng2020graph}, edge dropping \cite{bevilacqua2021equivariant,rong2020dropedge}, node marking \cite{papp2022theoretical}, and ego-networks \cite{zhao2022from,zhang2021nested,you2021identity}. The methods involving the removal of nodes or edges can also be viewed as regularization tools for GNNs, helping to reduce over-fitting \cite{feng2020graph,rong2020dropedge}. However, these methods often require a large number of perturbed graphs (or subgraphs) for each input graph, proportional to the graph size. For example, the number of perturbations that DropGNN \cite{papp2021dropgnn} requires is set to be the average node degree (e.g., 75 perturbed graphs for each graph in the $\collab$ dataset). Applying GNNs to this many perturbed graphs for a single input graph proves to be impractical, hindering scalability. Our approach shares some similarities with this body of research in generating perturbed graphs, albeit with a notable distinction: By once preprocessing diffused features of perturbed graphs, we ease the training computations, thus significantly improving scalability compared to DropGNN and other similar methods.
\vskip 0.5mm
\noindent\textbf{Feature augmented GNNs (Expressivity).} Several studies have aimed to increase the expressiveness of GNNs by enriching or augmenting node features. Some examples of auxiliary/augmented node information include geodesic distances  \cite{li2020distance,zhang2021labeling,velingker2022affinity}, resistance distances \cite{velingker2022affinity}, subgraph-induced structural information \cite{bouritsas2022improving,barcelo2021graph}, and positional encodings \cite{dwivedi2022graph}. % Detecting specific substructures within a graph (e.g., cycles with specific lengths) is important for various predictive tasks (e.g., computational chemistry \cite{koyuturk2004efficient}, social network analysis \cite{jiang2010finding}). Accordingly, certain approaches include substructure counts in the node features \cite{bouritsas2022improving, barcelo2021graph}. 
Despite the potential of feature augmentation to improve expressiveness, they fall short in scalability, as computation of the auxiliary features (e.g., substructure counts, the shortest distance between pairs, etc.) is expensive for large graphs.
\begin{figure*}[tb]
    \centering
    \begin{subfigure}[b]{0.42\textwidth}
        \centering
        \includegraphics[width=\textwidth]{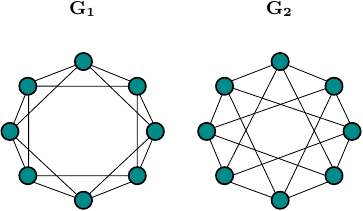}
        \caption{Two non-isomorphic graphs, indistinguishable by 1-WL test with the same label distributions.}
        \label{fig:sub2}
    \end{subfigure}
    \hspace{1.2cm}
    \begin{subfigure}[b]{0.42\textwidth}
        \centering
        \includegraphics[width=\textwidth]{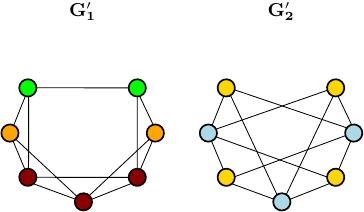}
        \caption{Perturbed input graphs, distinguishable by 1-WL test with different label distributions.}
        \label{fig:sub4}
    \end{subfigure}
    \caption{$1$-WL graph isomorphism test (a) fails to distinguish between two non-isomorphic graphs $G_1$ and $G_2$, but (b) successfully detect their perturbation (through node removal)  $G_1^{\prime}$ and $G_2^{\prime}$. }
    \label{fig:wl}
\end{figure*}
\section{Preliminary and Background}
We consider an undirected graph $G = (V, E)$ with $|V| = n$ nodes, $|E| = m$ edges, and adjacency matrix $\mathbf{A} \in \mathbb{R}^{{n}\times{n}}$. Each node $i \in V$ possesses the $d$-dimensional feature vector $\mathbf{x}_i \in \mathbb{R}^{n}$, which can be viewed as the $i$-th row of $n\times d$ feature matrix $\mathbf{X}$.

\subsection{Graph Prediction Task}
Graph classification or regression involves predicting a label (e.g., carcinogenicity classification \cite{toivonen2003statistical}) or a property (e.g., molecule solubility level \cite{gilmer2017neural%,lusci2013deep
}) for an entire graph based on its structure and associated features (e.g., node or edge features).
\footnote{Although our proposed models readily apply to other downstream tasks on graphs such as node classifications or link predictions, we focus on graph-level prediction tasks for conciseness.} Specifically, the task is formulated as a supervised learning problem, aiming to learn a mapping function $f:\mathcal{G} \to \mathcal{Y}$, given a labeled dataset $D=\{(G_i,y_i)\}$, where $\mathcal{G}$ is input space, and $\mathcal{Y}$ is class label space (or real for regression), $G_i$ is input graph sample, and $y_i$ is an expected label (or property). Graph Neural Networks have demonstrated significant success in effectively addressing the challenges of graph classification or regression tasks \cite{hamilton2020graph}.

\subsection{Graph Neural Networks}
In message-passing \textit{Graph Neural Networks (GNNs)}, each node's representation $\mathbf{h}_i$ is iteratively \textit{updated} and refined through the \textit{aggregation} of messages received from its neighbor's representations.\footnote{We use the abbreviation GNNs interchangeably with MPGNNs, although MPGNNs are a subclass of GNNs.} In GNNs, the node $i$'s representation at step (or layer) $l$ is updated by:
\begin{equation}
    \mathbf{h}_i^{(l)} = \upd\left(\mathbf{h}_i^{(l-1)}, 
    \mathbf{m}_{N(i)}^{(l-1)}\right)
\end{equation}    
where $\mathbf{m}_{N(i)}^{(l-1)}$ is the message aggregated from $i$'s neighborhood $N(i)$:
\begin{equation}   
    \mathbf{m}_{N(i)}^{(l-1)}   = \agg\left(\left\{\mathbf{h}_j^{(l-1)}\middle|j \in N(i)\right\}\right).
\end{equation}
GNNs usually differ from one another in how their $\agg$ and $\upd$ functions are defined. For example, \textit{graph convolutional networks (GCN)} \cite{kipf2016semi} employ a degree-normalized weighted mean as the $\agg$ function:
\begin{equation}
\mathbf{m}_{N(i)}^{(l-1)} = \sum_{j \in N(i)}\frac{\mathbf{h}_j^{(l-1)}}{\sqrt{|N(i)||N(j)|}}, 
\end{equation}
followed by a simple update function of:
\begin{equation}\mathbf{h}_i^{(l)}=\sigma\left(\mathbf{W}^{(l)}\left(\frac{\mathbf{h}_i^{(l-1)}}{|N(i)|}+ {\mathbf{m}}_{N(i)}^{(l-1)}\right)\right),
\end{equation}
%to Do 
where $\sigma$ is a non-linearity function (e.g., ReLU), and $\mathbf{W}^{(l)}$ is the $l^{th}$ layer’s learnable weight matrix. In GNNs, for each node $i\in V$, the initial node representation at step $0$ is usually set to their original node features: $\mathbf{h}_i^{(0)} = \mathbf{x}_i$. After $L$ message-passing iterations, the node $i$'s representation $\mathbf{z}_i$ is the output of layer $L$, i.e., $\mathbf{z}_i = \mathbf{h}_i^{(L)}$, or a combination of all $L$ layers' outputs $\mathbf{z}_i = \comb(\mathbf{h}_i^{(1)}, \dots ,\mathbf{h}_i^{(L)})$ \cite{xu2018representation}. One can view all learned node's representations in the form $n\times d'$ matrix $\mathbf{Z}$, where $d'$ is the hidden dimensionality. All nodes' final representations are then aggregated to form a graph (vector) representation:
\begin{equation}
    \mathbf{z}_G = \pool\left(\left\{\mathbf{z}_i \middle|i \in G\right\}\right),
\end{equation}
where the $\pool$ function can be non-adaptive (e.g., element-wise mean or max) or adaptive (e.g., top-k pooling \cite{lee2019self,gao2019graph}, set transformer \cite{buterez2022graph}, or MLP \cite{buterez2022graph}). The class probabilities are then computed by passing $\mathbf{z}_G$ through a non-linear learnable transformation (e.g., MLP). %The class probabilities $\mathbf{y}$ are computed by passing $\mathbf{z}_G$ through a learnable non-linear transformation $\mathbf{y} = \psi(\mathbf{z_G}\mathbf{W})$, where $\psi$ is a non-linear function (e.g., softmax or sigmoid), and $\mathbf{W}$ is a learnable weight matrix.

Despite the success of GNNs, their expressive power is limited and upper bounded by the $1$-WL test \cite{xu2018how,morris2019weisfeiler}. This limitation necessitates the development of new approaches with enhanced expressiveness.
\begin{figure*}[t]
  \centering
  \includegraphics[width=0.95\textwidth]{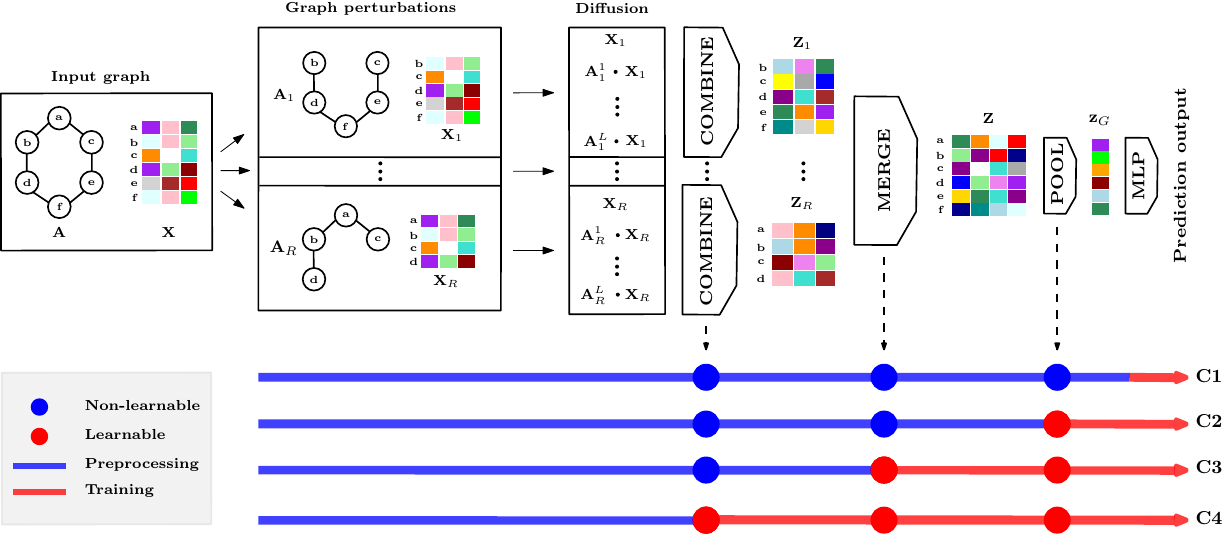}
  \caption{The \textit{SE2P} framework first generates $R$ perturbations of the input graph, where each perturbation involves randomly removing some nodes, thereby resulting in new adjacency and feature matrices ${(\mathbf{A}_r,\mathbf{X}_r)}$. Next, node features are diffused for each perturbation by a set of diffusion matrices: the perturbed adjacency matrix powers. Then, the $\comb$ function combines these diffused features for each perturbed graph to produce feature matrices $\mathbf{Z}_r$. All these matrices then undergo the $\merge$ function to generate a single nodal representation matrix $\mathbf{Z}$ for the input graph. Focusing on the graph-level task, we then apply $\pool$ to yield a graph (vector) representation $z_G$ and further achieve the predicted output through a non-linear transformation by MLP. The functions $\comb$, $\merge$, and $\pool$ can be either non-learnable (blue circle) or learnable (red circle). This flexibility allows us to choose between different configuration classes (C1, C2, C3, and C4) to balance scalability, achieved by including more preprocessing steps (blue line), and expressivity, achieved by having more learnable components and a longer training phase (red line).
  }
  \label{fig:SE2P}
  \vspace{-12pt}
\end{figure*}
\subsection{Perturbed Graph Neural Networks}
To overcome $1$-WL expressivity limitation of canonical GNNs \cite{xu2018how,morris2019weisfeiler}, \textit{Perturbed GNNs} (e.g., DropGNN \cite{papp2021dropgnn}) applies a shared GNN on $R$ different perturbations of the input graph (during both training and testing). For each perturbation $(\mathbf{A}_r, \mathbf{X}_r)$, some graph structure (e.g., nodes or edges) is randomly changed. For example, DropGNN randomly drops out some nodes for each perturbation, allowing a shared $L$-layer GNN to operate on a slightly perturbed version of the input graph to generate perturbed node representations $\mathbf{Z}_r = \gnn(\mathbf{A}_r, \mathbf{X}_r)$. These perturbed embeddings are then merged into final node embeddings using an aggregator function:
\begin{equation}
    \mathbf{Z} = \merge\left(\mathbf{Z}_1, \cdots,  \mathbf{Z}_R\right),
    \label{eq:node_z}
\end{equation}
where $\merge$ can be an element-wise mean operator. Through multiple perturbations, the model observes slightly perturbed variants of the same $L$-hop neighborhood around any node. Thus, even if the non-isomorphic neighborhoods are indistinguishable by the standard GNNs (or $1$-WL), their randomly modified variants are more likely to be distinguishable, yielding higher expressive power. For example, in Figure \ref{fig:wl}a, there are two non-isomorphic graphs which differ by the presence or absence of 3-length cycles (left graph vs right graph). The 1-WL algorithm cannot distinguish these two graphs as non-isomorphic. However, after a slight perturbation by removing a node (see Figure \ref{fig:wl}b), the 1-WL test successfully identifies them as non-isomorphic. 

For graph classification, the node representations $\mathbf{Z}$ in Equation \ref{eq:node_z} can be aggregated by a $\pool$ function into a graph-level representation $\mathbf{z}_G$. However, Perturbed GNNs (e.g., DropGNN) face major scalability issues as the number of perturbations increases, hindering their effectiveness on large datasets with high average node degrees.

\subsection{Simplified Diffusion-Based Models}
A practical approach to enhance the scalability of GNNs is simplifying their architectures by eliminating their intermediate non-linearities \cite{wu2019simplifying,sign_icml_grl2020,louis2023simplifying,zhang2022nafs,feng2020graph}. This simplification technique allows for the precomputation of feature propagation and further acceleration. For instance, %the \textit{Simplified Graph Convolutional Network (SGCN)} 
SGCN \cite{wu2019simplifying} removes intermediate non-linearities in an $L$-layer GCN, to predict node class labels $\mathbf{Y}$ using  $\mathbf{Y}=\sigma(\mathbf{A}^L\mathbf{XW})$, where $\sigma$ is a non-linear function, and $\mathbf{W}$ is a learnable weight matrix.
% where $\psi$ could be a softmax or logistic regression function, and $\mathbf{W}$ is a trainable weight matrix. 
The diffusion term $\mathbf{A}^L\mathbf{X}$ can be precomputed once before training. Extending upon SGCN, %\textit{Scalable Inception Graph Neural Networks (SIGN)}
SIGN \cite{sign_icml_grl2020} considers a set of diffusion matrices $\{\mathbf{S}^{(i)}\}$ (rather than just one diffusion matrix) to apply to the feature matrix $\mathbf{X}$:
\begin{equation}
\mathbf{Z}=\sigma\left(\bigoplus_{i}\mathbf{S}^{(i)}\mathbf{X}\mathbf{W}_i\right),
\end{equation}
where $\mathbf{W}_i$ is a learnable weight matrix associated with the diffusion matrix $\mathbf{S}^{(i)}$. Like SGCN, the terms $\mathbf{S}^{(i)}\mathbf{X}$ can be precomputed before training to speed up computation. %Applying a readout function to $\mathbf{Z}$ produces a graph-level representation $\mathbf{z}_G$ for graph classification or regression tasks. 
% The probabilities of class labels $y$ are then computed by passing this representation to another transformation, i.e., $\mathbf{y}=\psi(\mathbf{z}_G\mathbf{W}^\prime)$.
The diffusion terms in our model share some similarities to SIGN. However, unlike our method, the expressivity of SGCN and SIGN is bounded by the $1$-WL, as they were proposed to make the traditional GNNs more scalable.

\section{SE2P}
Inspired by the expressive power of methods relying on generating perturbations (e.g., DropGNN \cite{papp2021dropgnn}), yet motivated to address the scalability limitations, we propose \textit{\textbf{S}calable \textbf{E}xpressiveness through \textbf{P}reprocessed Graph \textbf{P}erturbations (SE2P)}. In SE2P, we first generate different perturbations of the input graph (e.g., through random node dropout) to improve expressiveness.\footnote{We explore node dropout as the perturbation type due to its theoretically-proved enhanced expressiveness \cite{papp2021dropgnn}. We hypothesize other types of perturbations would also offer expressive power but leave the theoretical and empirical exploration as future work.} The scalability is offered by once precomputing feature diffusions over perturbed graphs and eliminating the need for message-passing during training.

As illustrated in Fig.~\ref{fig:SE2P}, SE2P generates a set of $R$ graph perturbations $\{(\mathbf{A}_r,\mathbf{X}_r)\}$ for graph $G$ with adjacency matrix $\mathbf{A}$ and feature matrix $\mathbf{X}$. Although SE2P accommodates any perturbation kind (e.g., node removal, subgraph sampling, etc.), we here consider random nodal removal as a perturbation.  In each perturbation $(\mathbf{A}_r,\mathbf{X}_r)$, any node of the original graph $G$ is removed with probability $p$.\footnote{A common implementation trick for the removal of a node is to set its corresponding row and column in the adjacency matrix to all zeros. This trick allows the dimensionality of the adjacency matrices of all perturbed graphs to remain the same, thus easing aggregations downstream in pipelines. When this trick is used, the feature matrix for perturbation graphs is kept as the original one.} Each perturbed adjacency matrix $\mathbf{A}_r$ is normalized by $\hat{\mathbf{A}}_r = \mathbf{D}_r^{-\frac{1}{2}}\mathbf{A}_r\mathbf{D}_r^{-\frac{1}{2}}$, where $\mathbf{D}_r$ is the diagonal matrix of $\mathbf{A}_r$.\footnote{While adding self-loops to the adjacency matrix could offer certain benefits (e.g., over-fitting prevention), we deliberately avoid them in our approach. Our decision is backed up by the understanding that self-loops restrict generalizability by making the self or other's information indistinguishable  \cite{hamilton2020graph}.}
To emulate the message-passing of GNNs on perturbed graphs, we apply feature diffusion by $\hat{\mathbf{A}}_r\mathbf{X_r}$. Similarly, the message passing of an $L$-layer GNN can be emulated by $\hat{\mathbf{A}}_r^L\mathbf{X_r}$, which can be once precomputed before the training for each perturbed graph as a preprocessing step. For each perturbed graph, to have a more expressive node representation, we emulate the jumping knowledge \cite{xu2018how} by
\begin{equation}\label{eq:combine}
    \mathbf{Z}_r = \comb(\mathbf{X}, \hat{\mathbf{A}}_r^1\mathbf{X}_r,\cdots, \hat{\mathbf{A}}_r^L\mathbf{X}_r),
\end{equation}
where the $\comb$ function combines all the virtual $L$ layer's output with the original feature matrix into the node embedding matrix of the perturbed graph. The examples of $\comb$ can be simple readout-type operators (e.g., column-wise vector concatenation, etc.) or learnable adaptive aggregation mechanisms (e.g., LSTM or DeepSet). When the simple non-learnable operator is deployed, we compute $\mathbf{Z}_r$ through preprocessing steps for more speedup.
% For graphs with a high average degree, we select a lower $L$ to prevent over-smoothing \cite{zhang2022nafs}.

The next step is to aggregate node representations of perturbed graphs $\{\mathbf{Z}_r\}$ to a single nodal representation matrix $\mathbf{Z}$:
\begin{equation}
\label{eq:merge}
        \mathbf{Z}=\merge(\mathbf{Z}_1, \cdots, \mathbf{Z}_R),
\end{equation}
where several options exist for $\merge$ ranging from non-learnable aggregation methods (e.g., element-wise mean) to learnable set aggregations (e.g., DeepSet \cite{zaheer2017deep}).
While non-learnable aggregation methods, such as averaging, provide simplicity and computational efficiency, they may override and blend information across perturbed graphs, possibly leading to the loss or dilution of discriminative information. However, all computations up to this point can occur during the preprocessing phase, provided that aggregations in Eqs. \ref{eq:combine} and \ref{eq:merge} are non-learnable. This preprocessing offers a considerable speedup, as it emulates the message-passing of a multi-layer GNN on multiple perturbed graphs through a one-time preprocessing step rather than iterative computations during training. When more expressiveness is desired over scalability, one can employ learnable aggregation over perturbed graphs.

For graph prediction tasks (i.e., graph classification and regressions), we then apply a $\pool$ function to aggregate nodes' final representations into a graph (vector) representation:
\begin{equation}
     \mathbf{z}_G = \pool(\mathbf{Z}),
 \end{equation}
where $\mathbf{z}_G$ is the graph representation and the $\pool$ function can be non-learnable (e.g., element-wise sum) or a learnable graph pooling method. Non-learnable functions can speed up computation (specifically if they facilitate precomputation); However, they come at the cost of not introducing any non-linearity to the model to output graph representation, thus reducing the model's expressive power to some extent. If higher expressiveness is desired, given some computational budget, one might consider various learnable graph pooling methods such as hierarchical or top-k pooling \cite{ying2018hierarchical,gao2019graph,knyazev2019understanding,cangea2018towards}, global soft attention layer \cite{li2015gated}, set-transformer \cite{lee2019set}, or even MLP combined non-learnable aggregators (e.g., sum or mean). 
% In our setup, as we prioritize the scalability of the framework, we opt for a non-learnable pooling operator. Specifically, we use sum pooling as the $\readout$ function, as it offers greater expressive power compared to mean or max pooling \cite{xu2018how}. 
After the pooling operation, we apply a learnable non-linear function $\omega$ to get the class probabilities $\mathbf{y}$ from graph representation $\mathbf{z}_G$:
\begin{equation}
    \mathbf{y} = \omega(\mathbf{z}_G).
\end{equation}

\vskip 1mm 
\noindent \textbf{How does SE2P trade-off scalability and expressivity?} The SE2P's aggregation functions $\comb$, $\merge$, and $\pool$ offer a balance between scalability and expressivity. We can configure each to belong to either learnable or non-learnable aggregator classes. Thus, we specify four practical \textit{configuration classes} within SE2P for balancing scalability vs expressiveness. Each configuration class is identified by which aggregator is learnable or not.\footnote{Given three aggregation functions, and two choices of learnability for each, one can theoretically identify 8 potential classes. But, not all those classes would be beneficial for scalability if a learnable aggregator comes before a non-learnable one.}  All functions can be configured to be non-learnable (e.g., element-wise mean, sum, or max) for maximum scalability (see configuration class C1 in Fig. \ref{fig:SE2P}). This offers the highest scalability as all the computations up to obtaining the graph (vector) representation can be once precomputed before training. By exploiting a learnable function for $\pool$ while keeping $\comb$ and $\merge$ non-learnable, we can improve the expressivity of the model before generating the graph representation (see configuration class C2 in Fig.~\ref{fig:SE2P}). In this class, one can still preprocess the perturbed graphs, their diffusion, $\comb$, and $\merge$ for improved efficiency. Going one step further to enhance the expressivity, we can select $\merge$ as learnable in addition to $\pool$ (see configuration class C3 in Fig.~\ref{fig:SE2P}). In this class, the diffused matrices of the graph perturbations will be merged in the training stage. Given the desired computational resources, one can select learnable functions for all the aggregation functions. This configuration will be the least scalable configuration class (C4 in Fig. \ref{fig:SE2P}). Moving from C1 to C4 increases expressiveness but reduces scalability, as it allows less preprocessing to ease training’s computational burden.
\vskip 1mm 
\noindent \textbf{Our experimented configurations for SE2P.}
We have implemented and studied four instances of SE2P, covering all four configuration classes, with specific $\comb$, $\merge$, and $\pool$ functions.
$\confone$ uses column-wise vector concatenation for $\comb$, element-wise mean for $\merge$, and element-wise sum pooling for $\pool$. With all non-learnable aggregation functions, all operations up to generating the graph representations are precomputed for maximum scalability.
$\conftwo$ replaces the sum pooling of $\confone$ with a learnable $\pool$ function, which consists of an MLP followed by element-wise sum pooling.
$\confthree$ is the same as $\conftwo$ except for leveraging DeepSet as a learnable $\merge$ function (rather than non-learnable element-wise mean in $\conftwo$). The least scalable configuration $\conffour$ replaces the non-learnable $\comb$ of $\confthree$ to DeepSet, making all aggregator functions learnable for maximum expressivity. Table \ref{tab:configs} summarizes the selected aggregation functions of the configurations.

\begin{table}[tb]
\centering
\renewcommand{\arraystretch}{1.1}
\caption{The experimented \textcolor{blue}{non-learnable} or \textcolor{red}{learnable} aggregation functions of the SE2P configurations. }
\label{tab:configs}
\begin{tabular}{@{}l@{\hskip 0.7cm}c@{\hskip 0.7cm}c@{\hskip 0.7cm}c@{}}
\toprule
\textbf{Configuration} & \boldmath{$\comb$} & \boldmath{$\merge$} & \boldmath{$\pool$}\\
\midrule
$\confone$    & \textcolor{blue}{Concat}  & \textcolor{blue}{Mean}     &  \textcolor{blue}{Sum} \\
$\conftwo$    & \textcolor{blue}{Concat}  & \textcolor{blue}{Mean}     &  \textcolor{red}{MLP\ +\ Sum} \\
$\confthree$  & \textcolor{blue}{Concat}  & \textcolor{red}{DeepSet}   &  \textcolor{red}{MLP\ +\ Sum} \\
$\conffour$   & \textcolor{red}{DeepSet}  & \textcolor{red}{DeepSet}   &  \textcolor{red}{MLP\ +\ Sum} \\
\bottomrule
\end{tabular}
\vspace{-6pt}
\end{table}

%We have implemented and studied four instances of SE2P, with specific $\comb$, $\merge$, and $\pool$ functions.$\confone$ uses column-wise vector concatenation for the $\comb$ function to aggregate the virtual layers of the diffusion term for each perturbation, element-wise mean for the $\merge$ function to aggregate the feature matrices of the perturbations, and element-wise sum pooling for the $\pool$ function to generate the graph (vector) representation. This means that all operations up to generating the graph representations are precomputed.
%$\conftwo$ utilizes the same $\comb$ and $\merge$ functions as $\confone$. However, to introduce some non-linearity before generating the graph representations, we employ a learnable $\pool$ function, which consists of an MLP followed by element-wise sum pooling.
%$\confthree$ uses the non-learnable $\comb$ function same as the previous configuration, leveraging DeepSet as the $\merge$ function, and MLP followed by element-wise sum pooling as the $\pool$ function.
%$\conffour$, which is the least scalable configuration, considers learnable functions for all the aggregators. $\comb$ uses DeepSet to aggregate over the virtual layers of the diffusion terms for each perturbation, $\merge$ uses DeepSet to aggregate the feature matrices of the perturbations, and $\pool$ is the MLP-combined sum pooling to generate the graph representations.

\subsection{Runtime Analysis}
We investigate the (amortized) preprocessing and training time complexity per graph instance for DropGNN and our examined SE2P variants (see Table~\ref{tab:timecomplexity} for summary). For simplicity of representations, through our analyses, we assume that both input data dimensionality $d$ and hidden representation dimensionality $h$ have the same order of dimensionality $O(d)$.

For DropGNN, the preprocessing time is $O(1)$. For its training, the $R$ perturbations are generated in $O(Rn)$ by creating a dropout mask to zero out the connection of removed nodes, assuming that the graph is stored in a sparse matrix. For inference, the model applies $L$ layers of graph convolution and a two-layer MLP on each perturbation, which runs in $O(RL(nd^2+md) + nd^2)$ time, assuming $d$ is the feature dimension and proportional to the hidden dimensionality of both graph convolutions and MLP, and $m$ is number of edges. The inference time complexity of DropGNN is $O(Rn+RL(nd^2+md) + d^2) = O(RL(nd^2+md))$.

\begin{table}[h]
\centering
\renewcommand{\arraystretch}{1.1}
\caption{Running time complexities of DropGNN and SE2P configurations. $R$ denotes the number of perturbations, $L$ is the number of virtual layers of the diffusion step, $n$ is the number of nodes, $m$ is the number of edges, and $d$ is the feature dimensionality or the hidden dimensionality.}
\label{tab:timecomplexity}
\begin{tabular}{@{}l@{\hskip 0.8cm}l@{\hskip 0.8cm}l@{}}
\toprule
\textbf{Model} & \textbf{Preprocessing} & \textbf{Inference} \\
\midrule
DropGNN        & $O(1)$                 & $O(LR(nd^2+md))$ \\
$\confone$    & $O(RLmd)$              & $O(d^2)$ \\
$\conftwo$    & $O(RLmd)$              & $O(nd^2)$ \\
$\confthree$  & $O(RLmd)$              & $O(Rnd^2)$ \\
$\conffour$   & $O(RLmd)$               & $O(RLnd^2)$ \\
\bottomrule
\end{tabular}
\end{table}

\begin{table*}[ht]
\caption{Statistics of all the datasets used in our experiments.}
\label{tab:datasetstats}
\centering
\renewcommand{\arraystretch}{1.1} % Adjust spacing between rows
\begin{tabularx}{\textwidth}{@{}l@{\hspace{0.25cm}}l@{} *{5}{>{\centering\arraybackslash}X@{}}@{}}
\toprule
\textbf{Dataset} & \textbf{Task} & \textbf{\#Graphs} & \textbf{Avg. Nodes} & \textbf{Avg. Deg.} & \textbf{\#Classes} & \textbf{Feat. Type} \\
\midrule
MUTAG & Classify mutagenicity of compounds & $188$ & $17.93$ & $3.01$ & $2$ & Original\\
PROTEINS & Classify enzyme \& non-enzyme & $1109$ & $39.06$ & $5.78$ & $2$ & Original\\
PTC-MR & Classify chemical compounds & $344$ & $ 14.29$ & $3.18$ & $2$ & Original\\
IMDB-B & Classify movie genre & $996$ & $19.77$ & $18.69$ & $2$ & Encoded\\
IMDB-M &  Classify movie genre & $1498$ & $12.9$ & $11.91$ &  $3$ & Encoded\\
COLLAB & Classify researcher fields & $5000$ & $74.49$ & $73.49$ & $3$ & Encoded\\
\hline
OGBG-MOLTOX21 & Classify qualitative toxicity measurements & $7831$ & $19$ & $3.27$ & $12$ & Original\\
OGBG-MOLHIV & Classify HIV virus replication & $41127$ & $ 25.5$ & $3.38$ & $2$ & Original\\
\bottomrule
\end{tabularx}
\end{table*}

The preprocessing time for all SE2P variants is $O(RLmd)$. The preprocessing in $\conffour$ involves two parts: (i) creating $R$ node-removal perturbations in $O(Rn)$ time; (ii) diffusion of features over $RL$ sparse adjacency matrices of graphs with $O(m)$ edges in $O(RLmd)$ time. Note that $O(Rn + RLmd) = O(RLmd)$. The SE2P-C3 has an extra concatenation in $O(RLnd)$ time, dominated by $O(RLmd)$, thus again the total preprocessing time of $O(RLmd)$. The $\merge$ function of Mean in both $\confone$ and $\conftwo$ has a running time of $O(RLnd)$, and Sum for $\pool$ in $\confone$ has a running time of $O(Rnd)$; all are dominated by $O(RLmd)$, rendering the total running time for $\confone$ and $\conftwo$ again $O(RLmd)$.\footnote{To achieve faster and more scalable preprocessing implementations for SE2P, one can parallelize sparse matrix multiplications using frameworks for distributed computing (e.g., Spark, etc.), which is not explored in this work.}

For inference, $\confone$ only applies the MLP transformation to the graph representation for graph classification, resulting in a time complexity of $O(d^2)$. $\conftwo$ adds another learnable component in the $\pool$ function (MLP + SUM), with time complexity of $O(nd^2)$. Thus, the $\conftwo$ infer in $O(nd^2 + d^2) = O(nd^2)$ time. Compared to $\conftwo$, 
$\confthree$ has an additional learnable DeepSet for $\merge$ to combine perturbations. This DeepSet runs in $O(Rnd^2 + nd^2) = O(Rnd^2)$, dominating other running times of $\pool$ and final transformation, thus making $\confthree$ running time $O(Rnd^2)$. $\conffour$ has an additional DeepSet as $\comb$ for aggregating the $L$ virtual layer's representations in each perturbation, which runs in $O(RLnd^2)$. As this dominates other components' running time asymptotically, the time complexity of $\conffour$ is $O(RLnd^2)$.

\section{Experiments}
Our experiments aim to empirically validate the scalability (e.g., runtime, memory requirements, etc.) and generalizability of our SE2P models against various benchmarks on graph prediction tasks.
\vskip 1mm
\noindent \textbf{Datasets.} For graph classification tasks, we experiment with six datasets from the TUDataset collection \cite{Morris+2020}: three \textit{bioinformatics datasets} of $\mutag$ \cite{debnath1991structure}, $\proteins$ \cite{borgwardt2005protein,dobson}, and $\ptc$ \cite{toivonen2003statistical}; and three \textit{social network datasets} of $\imdbb$ \cite{yanardag2015deep}, $\imdbm$ \cite{yanardag2015deep}, and $\collab$ \cite{yanardag2015deep}. These datasets have been the subject of study for other prominent graph classification methods \cite{xu2018how,bevilacqua2021equivariant,frasca2022understanding,papp2021dropgnn,zhao2022from}. The bioinformatics datasets contain categorical nodal features, while the social network datasets do not have any nodal features. As with others \cite{xu2018how,bevilacqua2021equivariant,papp2021dropgnn}, we encode node degrees as node features for datasets without nodal features. For TU datasets, we use the same dataset splitting deployed by other related studies \cite{xu2018how,bevilacqua2021equivariant,papp2021dropgnn,zhao2022from}, which is 10-fold cross-validation. We also test our model on the $\ogbmolhiv$ and $\ogbmoltox$ datasets from Open Graph Benchmark (OGB) \cite{hu2020open} using their provided scaffold splits. Their provided splits have been widely used among other works evaluating OGB benchmarks \cite{bevilacqua2021equivariant,zhao2022from}. We provide the statistics of all datasets in Table \ref{tab:datasetstats}.
\vskip 1mm
\noindent \textbf{Baselines.}
For graph classification on TU datasets, we assess and compare our model with several state-of-the-art baselines: the WL subtree kernel \cite{shervashidze2011weisfeiler}, Diffusion convolutional neural networks (DCNN) \cite{atwood2016diffusion}, Deep Graph CNN (DGCNN) \cite{zhang2018end}, PATCHY-SAN \cite{niepert2016learning}, IGN \cite{maron2018invariant}, GCN \cite{kipf2016semi}, GIN \cite{xu2018how}, DropGIN \cite{papp2021dropgnn}, and DropGCN.\footnote{Our introduced DropGCN has replaced GIN layers with GCN layers in DropGIN.}
% We present the accuracies of these benchmarks as reported in their original papers. Moreover, we have reproduced the results for some baselines, including GCN \cite{kipf2016semi}, GIN \cite{xu2018how}, DropGIN \cite{papp2021dropgnn}, and our implementation of DropGCN, a modified version of DropGIN.
On open Graph Benchmark datasets, we compare our proposed models against GCN, GIN, DropGCN, and DropGIN.
\begin{table*}[t]
\caption{Average validation accuracy (\%), TU datasets. The best result is in bold. In parenthesis: the ranks of our model against baselines (\textcolor{customgreen}{\textbf{1}}st, \textcolor{customyellow}{\textbf{2}}nd, and \textcolor{customred}{\textbf{3}}rd are colored), and comparison to DropGNNs ($\greencircle$=better, $\yellowcircle$=comparable with difference < 0.2, and $\redcircle$=worse). OOM denotes out of memory.}
\label{tab:TUresults}
\centering
%\scriptsize
\renewcommand{\arraystretch}{1.1} % Adjust spacing between rows
%\begin{tabularx}{1.0\textwidth}{@{}l@{\hskip 0.00cm} *{6}{>{\centering\arraybackslash}X@{}}@{}}
\begin{threeparttable}
\begin{tabularx}{1.0\textwidth}{@{}l@{\hskip 1cm} *{6}{>{\raggedright\arraybackslash}X}@{}}
\toprule
\textbf{Model\ /\ Dataset} & \textbf{MUTAG} & \textbf{PROTEINS} & \textbf{PTC-MR} & \textbf{IMDB-B} & \textbf{IMDB-M} & \textbf{COLLAB}\\
\midrule
WL subtree & \textbf{90.4} $\pm$ \textbf{5.7} & $75.0 \pm 3.1$ & $59.9 \pm 4.3$ & $73.8 \pm 3.9$ & $50.9 \pm 3.8$ & $78.9 \pm 1.9$\\
DCNN & $67.0 \pm 2.1$ & $61.3 \pm 1.6$ & $56.6 \pm 1.2$ & $49.1 \pm 1.4$ & $33.5 \pm 1.4$ & $52.1 \pm 2.7$\\
DGCNN & $85.8 \pm 1.7$ & $75.5 \pm 0.9$ & $ 58.6 \pm 2.5$ & $ 70.0 \pm 0.9$ & $47.8 \pm 0.9$ & $73.7 \pm 0.4$\\
PATCHYSAN & $89.0 \pm 4.4$ & $75.0 \pm 2.5$ & $62.3 \pm 5.7$ & $71.0 \pm 2.3$ & $45.2 \pm 2.8$ & $72.6 \pm 2.2$\\
IGN &  $83.9 \pm 13.0$ & $76.6 \pm 5.5$ & $58.5 \pm 6.9$ & $72.0 \pm 5.5$ &  $48.7 \pm 3.4$ & $78.3 \pm 2.5$\\
GIN & $88.8 \pm 5.7$ & $75.4 \pm 5.0$ & $63.9 \pm 8.3$ & $75.5 \pm 4.0$ & $51.5 \pm 4.0$ & $82.2 \pm 2.1$\\
GCN & $85.5 \pm 9.4$ & $75.9 \pm 5.5$ & $64.2 \pm 9.7$ & $75.7 \pm 3.6$ & $52.0 \pm 4.1$ & $82.6 \pm 2.2$ \\
DropGIN & $89.1 \pm 9.2$  & $76.1 \pm 5.1$ & $65.2 \pm 9.8$ & $75.2 \pm 3.1$ & $52.3 \pm 3.8$ & OOM\tnote{a}\\
DropGCN & $87.1 \pm 9.7$ & $76.1 \pm 5.8$ & $64.5 \pm 9.1$ & $75.3 \pm 3.3$ & $52.1\pm 3.3$ & OOM\tnote{a}\\
\hline
$\confone$ & $87.8 \pm 6.9\ (5\ ,\redcircle)$ & $74.7 \pm 5.7\ (9\ ,\redcircle)$ & $64.5 \pm 8.0\ (\textcolor{customyellow}{\mathbf{2}}\ ,\redcircle)$ & $74.6 \pm 3.7\ (5\ ,\redcircle)$ & $52.1 \pm 2.8\ (\textcolor{customyellow}{\mathbf{2}}\ ,\yellowcircle)$ & $79.8 \pm 1.5\ (\textcolor{customred}{\mathbf{3}}\ ,\greencircle)$\\
$\conftwo$ & $89.4 \pm 7.4\ (\textcolor{customyellow}{\mathbf{2}}\ ,\greencircle)$ & $77.6 \pm 6.3\ (\textcolor{customgreen}{\mathbf{1}}\ ,\greencircle)$ & $65.1 \pm 7.3\ (\textcolor{customyellow}{\mathbf{2}}\ ,\yellowcircle)$ & $75.2 \pm 2.9\ (\textcolor{customred}{\mathbf{3}}\ ,\yellowcircle)$ & $52.3 \pm 2.3\ (\textcolor{customgreen}{\mathbf{1}}\ ,\yellowcircle)$ & $83.3 \pm 2.1\ (\textcolor{customgreen}{\mathbf{1}}\ ,\greencircle)$\\
$\confthree$ & $89.5 \pm 6.6\ (\textcolor{customyellow}{\mathbf{2}}\ ,\greencircle)$ & $\textbf{77.6} \pm \textbf{5.0}\  (\textcolor{customgreen}{\mathbf{1}}\ ,\greencircle)$ & $\mathbf{66.2 \pm 6.8}\ (\textcolor{customgreen}{\mathbf{1}}\ ,\greencircle)$ & $\textbf{75.9} \pm \textbf{3.4}\ (\textcolor{customgreen}{\mathbf{1}}\ ,\greencircle)$ & $\textbf{52.9} \pm \textbf{3.5}\ (\textcolor{customgreen}{\mathbf{1}}\ ,\greencircle)$ & $\textbf{83.5} \pm \textbf{1.7}\ (\textcolor{customgreen}{\mathbf{1}}\ ,\greencircle)$\\
$\conffour$ & $89.7 \pm 7.5\ (\textcolor{customyellow}{\mathbf{2}}\ ,\greencircle)$ & $76.8 \pm 4.7\ (\textcolor{customgreen}{\mathbf{1}}\ ,\greencircle)$ & $66.1 \pm 8.8\ (\textcolor{customgreen}{\mathbf{1}}\ ,\greencircle)$ & $75.7 \pm 4.8\ (\textcolor{customyellow}{\mathbf{2}}\ ,\greencircle)$ & $52.4 \pm 2.4\ (\textcolor{customgreen}{\mathbf{1}}\ ,\yellowcircle)$ & $82.8 \pm 2.1\ (\textcolor{customgreen}{\mathbf{1}}\ ,\greencircle)$\\
\bottomrule
\end{tabularx}
\begin{tablenotes}
\item[a] We encountered out-of-memory issues with DropGNN on the COLLAB dataset due to the large number of graphs and perturbations, but reducing the batch size (e.g., from 32 to 6 in DropGCN on our server) allows these models to run. However, this resulted in suboptimal performance, leading us to report OOM results to highlight computational bottlenecks rather than expressiveness concerns.
\end{tablenotes}
\end{threeparttable}
\end{table*}

\vskip 1mm
\noindent \textbf{Experimental setup.}
%For SE2P, we must determine the optimal values for several hyperparameters, such as the probability of dropping a node $p$, the number of perturbations $R$, and the number of virtual layers $L$ in Equation~\ref{eq:combine}. 
% We have implemented an instance of SE2P by utilizing column-wise vector concatenation as the $\comb$ function, an element-wise mean as the $\merge$ function, and MLP followed by element-wise sum pooling as the $\pool$ function. Our implemented $\conftwo$ falls into configuration class 2 with non-learnable functions for $\comb$ and $\merge$, and a learnable function for $\pool$. Consequently, all computations up to the $\merge$ function take place during the preprocessing step (see  C2 in Fig.\ref{fig:SE2P}).\footnote{The Github link of the project is omitted to preserve anonymity.}
For a fair comparison between DropGNN \cite{papp2021dropgnn} and SE2P variants, we adopt the recommended hyperparameters for DropGNN by setting the probability of dropping node $p = \frac{2}{1+\gamma}$ and the number of perturbations $R = \left\lfloor \gamma \right\rfloor$, where $\gamma$ is the average node degree in the dataset.\footnote{Theoretical analysis for choosing the number of perturbations have been provided in DropGNN \cite{papp2021dropgnn}.} The number of (virtual) layers $L$ is set to $2$ or $3$ for each dataset.\footnote{Motivated by the observation that a lower $L$ for denser graphs yield better generalization \cite{zhang2022nafs}, we have set L=2 if the average degree less than 10; otherwise L=3.} For TU benchmark evaluations, we present the accuracies of the WL subtree kernel, DCNN, DGCNN, PATCHY-SAN, and IGN, as reported in their original papers.\footnote{Similarly, many other relevant studies \cite{xu2018how,papp2021dropgnn,bevilacqua2021equivariant}, which share the same experimental setup as ours, have borrowed these results from their original papers.} Under this experimental setup, we also reproduced the results for GCN \cite{kipf2016semi}, GIN \cite{xu2018how}, DropGIN \cite{papp2021dropgnn}, and DropGCN. We grid-searched the hyperparameters for these baselines on the recommended search spaces \cite{papp2021dropgnn}. For the SE2P variants, we performed a hyperparameter search for each dataset using a cascading approach from $\confone$ to $\conffour$: Once hyperparameters are determined for a component of a model (e.g., final MLP transformation component in $\confone$), they are kept the same across all subsequent configurations (e.g., $\conftwo$ to $\conffour$). A subsequent hyperparameter search is then conducted on the immediate subsequent variant (e.g., $\conftwo$) while focusing solely on its introduced hyperparameters (e.g., hyperparameters for the $\pool$ function in $\conftwo$). Then, the tuned hyperparameters are again fixed and copied to all subsequent variants (e.g., $\confthree$ and $\conffour$). The cascading process continues until all hyperparameters of the last variant are tuned.\footnote{We note that our cascading hyperparameter tunning procedure is suboptimal, but this process considerably speeds up our hyperparameter search by limiting the exponential growth of search space and running the (sub)optimal hyperparameters of previous models.}

Similar to \textit{Xu et al.} \cite{xu2018how}, our experimental setup involves a $10$-fold cross-validation. For each fold, we optimized a model with Adam for $350$ epochs, and an initial learning rate of $0.01$ decayed by a factor of $0.5$ every $50$ epochs. In this setup, datasets are split into training and validation sets (without a separate test set), and the reported results are on the validation accuracy. We obtain 10 validation curves corresponding to the 10 folds, calculate the average validation curve across all folds, and then select an epoch that achieves the highest averaged validation accuracy. We also compute standard deviation over the 10 folds at the selected epoch. For the OGB benchmark, we employed the same hyperparameter tuning approach as we did for the TU benchmark, and then followed the evaluation procedure proposed in \cite{hu2020open}: we ran each experiment with 10 different random seeds, optimized using Adam for 100 epochs. We obtained the element-wise mean of the validation curves of all the seeds and determined the test accuracies (and their average) corresponding to the best validation accuracy.\footnote{The code is available at https://github.com/Danial-sb/SE2P. All experiments
were conducted on a server with 40 CPU cores, 377 GB RAM, and 11 GB GTX 1080 Ti GPU.}
\begin{table*}[ht]
\centering
%\small
\setlength{\tabcolsep}{1.52pt}
\renewcommand{\arraystretch}{1.1}
\caption{Runtime Analysis over TU datasets. The training time is the average time per epoch (over 350 epochs), and the runtime is the processing time and the training time. SE2Ps are color-coded by \colorbox{green!25}{faster},\colorbox{yellow!35}{comparable}, and  \colorbox{red!25}{slower} than the baselines. The maximum and minimum speedup corresponds to the ratio of time taken by the slowest and the fastest baselines compared to our model. All results are reported in seconds.}
\label{tab:runtimetu}
\adjustbox{max width=\textwidth}{%
\begin{tabularx}{1.0\textwidth}{@{}lcccccccccccccccccc@{}}
\toprule
& \multicolumn{3}{c}{\textbf{MUTAG}} & \multicolumn{3}{c}{\textbf{PROTEINS}} &\multicolumn{3}{c}{\textbf{PTC-MR}}& \multicolumn{3}{c}{\textbf{IMDB-B}} & \multicolumn{3}{c}{\textbf{IMDB-M}} & \multicolumn{3}{c}{\textbf{COLLAB}} \\
\cmidrule(lr){2-4} \cmidrule(lr){5-7} \cmidrule(lr){8-10}
\cmidrule(lr){11-13} \cmidrule(lr){14-16} \cmidrule(lr){17-19}
\textbf{Model}   & \textbf{Prep.} & \textbf{Train} & \textbf{Run} & \textbf{Prep.} & \textbf{Train} & \textbf{Run} & \textbf{Prep.} & \textbf{Train} & \textbf{Run} & \textbf{Prep.} & \textbf{Train} & \textbf{Run} & \textbf{Prep.} & \textbf{Train} & \textbf{Run} & \textbf{Prep.} & \textbf{Train} & \textbf{Run} \\
\cmidrule(r){1-1}\cmidrule(lr){2-4} \cmidrule(lr){5-7} \cmidrule(lr){8-10} \cmidrule(lr){11-13} \cmidrule(lr){14-16} \cmidrule(lr){17-19}
GIN & $-$ & $0.71$ & $248$ & $-$ & $ 0.74$ & $259$ & $-$ & $ 1.52$ & $532$ & $-$ & $ 0.71$ & $248$ & $-$ & $ 0.72$ & $252$ & $-$ & $3.87$ & $1354$\\
GCN & $-$ & $0.76$ & $266$ & $-$ & $ 0.71$ & $248$ & $-$ &  $0.64$ & $224$ & $-$ & $ 0.77$ & $269$ & $-$ & $0.73$ & $255$ & $-$ & $2.62$ & $918$\\
DropGIN & $-$ & $0.82$ & $287$ & $-$ & $ 0.86$ & $301$ & $-$ &  $ 1.74$ & $609$ & $-$ & $1.05$ & $367$ & $-$ & $ 0.94$ & $329$ & $-$ & OOM & OOM\\
DropGCN & $-$ & $ 1.04$ & $364$ & $-$ & $0.94$ & $329$ & $-$ &  $0.86$ & $303$ & $-$ & $0.99$ & $348$ & $-$ & $ 1.19$ & $418$ & $-$ &  OOM & OOM\\
\midrule
$\confone$ & $0.5$ & \colorbox{green!25}{$0.26$} & \colorbox{green!25}{$92$} & $8.7$ & \colorbox{green!25}{$0.27$} & \colorbox{green!25}{$103$} & $1.8$ & \colorbox{green!25}{$0.22$} & \colorbox{green!25}{$79$} & $3.8$ & \colorbox{green!25}{$0.22$} & \colorbox{green!25}{$81$} & $4.0$ & \colorbox{green!25}{$0.23$} & \colorbox{green!25}{$84$} & $230.3$ & \colorbox{green!25}{$0.26$} & \colorbox{green!25}{$321$}\\
Max speedup & $-$ & $4.00$ & $3.96$ & $-$ & $3.48$ & $3.19$ & $-$ & $7.90$ & $7.70$ & $-$ & $4.77$ & $4.53$ & $-$ & $5.17$ & $4.97$ & $-$ & $14.88$ & $4.21$\\
Min speedup & $-$ & $2.73$ & $2.69$ & $-$ & $2.62$ & $2.40$ & $-$ & $2.90$ & $2.83$ & $-$ & $3.22$ & $3.06$ & $-$ & $3.13$ & $3.00$ & $-$ & $10.07$ & $2.85$\\
\hline
$\conftwo$ & $0.5$ & \colorbox{green!25}{$0.42$} & \colorbox{green!25}{$148$} & $8.7$ & \colorbox{green!25}{$0.41$} & \colorbox{green!25}{$152$} & $1.7$ & \colorbox{green!25}{$0.28$} & \colorbox{green!25}{$101$} & $3.8$ & \colorbox{green!25}{$0.38$} & \colorbox{green!25}{$137$} & $4.0$ & \colorbox{green!25}{$0.37$} & \colorbox{green!25}{$133$} & $224.2$ & \colorbox{green!25}{$0.72$} & \colorbox{green!25}{$476$}\\
Max speedup & $-$ & $2.47$ & $2.45$ & $-$ & $2.29$ & $2.16$ & $-$ & $6.21$ & $6.02$ & $-$ & $2.76$ & $2.67$ & $-$ & $3.13$ & $3.14$ & $-$ & $5.37$ & $2.84$\\
Min speedup & $-$ & $1.69$ & $1.67$ & $-$ & $1.73$ & $1.63$ & $-$ & $2.28$ & $2.21$ & $-$ & $1.86$ & $1.81$ & $-$ & $1.89$ & $1.89$ & $-$ & $3.63$ & $1.92$\\
\hline
$\confthree$ & $0.5$ & \colorbox{yellow!35}{$1.19$} & \colorbox{yellow!35}{$417$} & $8.5$ & \colorbox{yellow!35}{$1.34$} & \colorbox{yellow!35}{$477$} & $1.6$ & \colorbox{yellow!35}{$0.70$} & \colorbox{yellow!35}{$247$} & $2.9$ & \colorbox{yellow!35}{$0.85$} & \colorbox{yellow!35}{$300$} & $3.1$ & \colorbox{yellow!35}{$0.82$} & \colorbox{yellow!35}{$291$} & $220.2$ & \colorbox{red!25}{$20.06$} & \colorbox{red!25}{$7241$}\\
Max speedup & $-$ & $0.87$ & $0.87$ & $-$ & $0.70$ & $0.68$ & $-$ & $2.48$ & $2.46$ & $-$ & $1.19$ & $1.22$ & $-$ & $1.45$ & $1.43$ & $-$ & $0.19$ & $0.18$\\
Min speedup & $-$ & $0.59$ & $0.59$ & $-$ & $0.52$ & $0.51$ & $-$ & $0.91$ & $0.90$ & $-$ & $0.83$ & $0.82$ & $-$ & $0.87$ & $0.86$ & $-$ & $0.12$ & $0.12$\\
\hline
$\conffour$ & $0.5$ & \colorbox{red!25}{$5.22$} & \colorbox{red!25}{$1827$} & $8.5$ & \colorbox{red!25}{$7.06$} & \colorbox{red!25}{$2479$} & $1.4$ & \colorbox{red!25}{$3.58$} & \colorbox{red!25}{$1254$} & $2.9$ & \colorbox{red!25}{$3.90$} & \colorbox{red!25}{$1368$} & $3.0$ & \colorbox{red!25}{$2.61$} & \colorbox{red!25}{$918$} & $214.7$ & \colorbox{red!25}{$42.86$} & \colorbox{red!25}{$15215$}\\
Max speedup & $-$ & $0.19$ & $0.19$ & $-$ & $0.13$ & $0.13$ & $-$ & $0.48$ & $0.48$ & $-$ & $0.26$ & $0.26$ & $-$ & $0.45$ & $0.45$ & $-$ & $0.09$ & $0.08$\\
Min speedup & $-$ & $0.13$ & $0.13$ & $-$ & $0.10$ & $0.09$ & $-$ & $0.17$ & $0.17$ & $-$ & $0.18$ & $0.18$ & $-$ & $0.27$ & $0.27$ & $-$ & $0.06$ & $0.05$\\
\bottomrule
\end{tabularx}%
}
\end{table*}
\begin{table*}[ht]
\centering
%\scriptsize
\caption{Average ROC-AUC (\%) over 10 runs on OGB datasets. The best is in bold. The preprocessing time, training time per epoch, and the total runtime are provided in seconds. Color-coding is \colorbox{green!25}{faster},\colorbox{yellow!35}{comparable}, and \colorbox{red!25}{slower} than any of baselines.}
\adjustbox{max width=\textwidth}{%
\renewcommand{\arraystretch}{1.15}
\label{tab:ogb}
\begin{tabularx}{\textwidth}{@{}l*{10}{>{\centering\arraybackslash}X@{\hskip 0.24cm}}@{}}
\toprule
\multirow{2}{*}{\textbf{Model}} & \multicolumn{5}{c}{\textbf{OGBG-MOLHIV}} & \multicolumn{5}{c}{\textbf{OGBG-MOLTOX21}} \\
\cmidrule(lr){2-6} \cmidrule(lr){7-11}
& \textbf{Validation} & \textbf{Test} & \textbf{Prep.} & \textbf{Time/epoch} & \textbf{Run} &\textbf{Validation} & \textbf{Test} &  \textbf{Prep.} & \textbf{Time/epoch} & \textbf{Run} \\
\midrule
GIN & $78.1 \pm 2.0$ & $74.0 \pm 1.9$ & $-$ & $3.54$ & $354$ & $75.6 \pm 1.1$ & $72.7 \pm 1.7$ & $-$ & $1.56$ & $156$ \\
GCN & \textbf{78.6} $\pm$ \textbf{2.4} & $74.1 \pm 1.9$ & $-$ & $3.56$ & $356$ & $76.0 \pm 0.8$ & $72.2 \pm 1.1$ & $-$ & $1.71$ & $171$ \\
DropGIN & OOM & OOM & $-$ & $-$ & $-$ & $75.7 \pm 0.7$ & $73.6 \pm 1.0$ & $-$ & $2.33$ & $233$\\
DropGCN & OOM & OOM & $-$ & $-$ & $-$ & $76.2 \pm 0.7$ & $72.1 \pm 1.1$ & $-$ & $2.52$ & $252$\\
\hline
$\confone$ & $73.1 \pm 2.0$ & $71.4 \pm 1.3$ & $170.5$ & \colorbox{green!25}{$1.79$} & \colorbox{yellow!35}{$349$} & $72.6 \pm 0.4$ & $71.6 \pm 0.5$ & $22.6$ & \colorbox{green!25}{$0.59$} & \colorbox{green!25}{$81$} \\
$\conftwo$ & $77.1 \pm 0.9$ & $74.0 \pm 1.3$ & $169.7$ & \colorbox{green!25}{$2.23$} & \colorbox{yellow!35}{$393$} & $76.1 \pm 0.8$ & $72.9 \pm 0.6$ & $22.5$ & \colorbox{green!25}{$0.87$} & \colorbox{green!25}{$109$}\\
$\confthree$ & $76.3 \pm 2.2$ & \textbf{74.5} $\pm$ \textbf{2.6} & $171.6$ & \colorbox{red!25}{$11.21$} & \colorbox{red!25}{$1292$} & $76.7 \pm 0.9$ & $73.5 \pm 1.0$ & $22.6$ & \colorbox{yellow!35}{$2.46$} & \colorbox{yellow!35}{$268$} \\
$\conffour$ & OOM & OOM & $173.1$ & $-$ & $-$ & \textbf{77.0} $\pm$ \textbf{1.3} & \textbf{74.1} $\pm$ \textbf{1.0} & $22.6$ & \colorbox{red!25}{$11.36$} & \colorbox{red!25}{$1158$}\\
% $\SE2Pds$ & OOM & OOM & - & $0.7606\pm0.0126$ & $0.7356\pm0.0076$ & $2.56 \pm 0.42$ \\
\bottomrule
\end{tabularx}}
\end{table*}

\vskip 1mm
\noindent \textbf{Results and Discussions.}
We present the validation accuracy results on TU datasets in Table~\ref{tab:TUresults}. In all datasets except $\mutag$, $\confthree$ outperforms other SE2P configurations and baselines, and improves the generalizability over all baselines ranging from 0.6\% (in $\imdbm$) to 1.5\% (in $\proteins$). $\conftwo$ and $\conffour$ also show competitive performance, securing the top three-ranked methods among all baselines for all datasets. For instance, $\conffour$ improves or shows comparable results to all baselines in all datasets except $\mutag$, where the WL subtree model performed best. Our least expressive $\confone$ model performs suboptimally on three datasets of $\mutag$, $\proteins$, and $\imdbb$, but is relatively competitive in other datasets (e.g., $\ptc$, $\imdbm$, $\collab$) by being ranked among the top three of baselines. The poor performance in those three datasets might be due to the lack of non-linearity before obtaining the graph representation and complexity of those datasets. Notably, for most datasets, models involving perturbation generation (SE2P, DropGCN, and DropGIN) outperform most baselines that do not utilize graph perturbation (e.g., GCN, IGN). This result suggests that graph perturbations are not only a theoretically-founded method for going above 1-WL expressiveness, but are a simple yet effective method for improving generalization. 
Comparing our SE2P models with DropGNN, which benefits from the power of graph perturbations, our configurations (except for $\confone$) always show comparable or improved generalizability. This generalization improvement is also complemented by handling the scalability issues of DropGNN (e.g., OOM in $\collab$ and longer training times for other datasets).

% in three datasets of {\footnotesize PROTEINS}, {\footnotesize PTC-MR}, and {\footnotesize COLLAB}. In three other datasets, our model achieves comparable results to the best baseline and ranked in the top 3. For most datasets, $\conftwo$ and DropGNNs (e.g., DropGCN and DropGIN) outperform most other baselines without node dropout (e.g., GCN, IGN). %This is primarily because, unlike other baselines that do not include generating graph perturbations, the inclusion of graph perturbations enhances the generalizability of SE2P.
\begin{table}[tb]
\centering
\renewcommand{\arraystretch}{1.15}
\caption{Ablation study comparing the impact of removing perturbations and the $\merge$ function (SIGN), removing perturbations and $\comb$ and $\merge$ functions (SGCN) versus $\conftwo$ on TU datasets. The preprocessing time (Pre), training time per epoch (Train), and the total runtime (Run) are provided in seconds. Reported results are in terms of graph classification validation accuracy (Val). The best is highlighted in bold.}
\vspace{-5pt}
\setlength{\tabcolsep}{1.4pt}
\label{tab:ablation}
\begin{tabularx}{1.0\columnwidth}{@{}lcccccccc}
\toprule
& \multicolumn{4}{c}{\textbf{\mutag}} & \multicolumn{4}{c}{\textbf{\proteins}} \\
\cmidrule(lr){2-5} \cmidrule(lr){6-9}
\textbf{Model} & \textbf{Pre} & \textbf{Train} & \textbf{Run} & \textbf{Val} & \textbf{Pre} & \textbf{Train} & \textbf{Run} & \textbf{Val} \\
\cmidrule(r){1-1}\cmidrule(lr){2-5} \cmidrule(lr){6-9}
SGCN & 0.1 & 0.41 & 144 & 87.7 $\pm$ 8.9 & 0.5 & 0.41 & 143 & 76.2 $\pm$ 5.3\\
%\hline
SIGN & 0.1 & 0.43 & 150 & 87.2 $\pm$ 9.6 & 0.7 & 0.41 & 143 & 76.4 $\pm$ 5.7\\
%\hline
$\conftwo$ & 0.5 & 0.42 & 148 & \textbf{89.4} $\pm$ \textbf{7.4} & 8.7 & 0.41 & 152 & \textbf{77.6} $\pm$ \textbf{6.3}\\
%\bottomrule
\toprule
& \multicolumn{4}{c}{\textbf{\ptc}} & \multicolumn{4}{c}{\textbf{\imdbb}} \\
\cmidrule(lr){2-5} \cmidrule(lr){6-9}
\textbf{Model} & \textbf{Pre} & \textbf{Train} & \textbf{Run} & \textbf{Val} & \textbf{Pre} & \textbf{Train} & \textbf{Run} & \textbf{Val} \\
\cmidrule(r){1-1}\cmidrule(lr){2-5} \cmidrule(lr){6-9}
SGCN & 0.1 & 0.29 & 102 & 63.6 $\pm$ 9.7 & 0.4 & 0.33 & 116 & 74.2 $\pm$ 3.4 \\
%\hline
SIGN & 0.2 & 0.30 & 105 & 64.5 $\pm$ 5.8 & 0.5 & 0.36 & 126 & 74.5 $\pm$ 5.4\\
%\hline
$\conftwo$ & 1.7 & 0.28 & 101 & \textbf{65.1} $\pm$ \textbf{7.3} & 3.8 & 0.38 & 137 & \textbf{75.2} $\pm$ \textbf{2.9}\\
%\bottomrule
\toprule
& \multicolumn{4}{c}{\textbf{\imdbm}} & \multicolumn{4}{c}{\textbf{\collab}} \\
\cmidrule(lr){2-5} \cmidrule(lr){6-9}
\textbf{Model} & \textbf{Pre} & \textbf{Train} & \textbf{Run} & \textbf{Val} & \textbf{Pre} & \textbf{Train} & \textbf{Run} & \textbf{Val} \\
\cmidrule(r){1-1}\cmidrule(lr){2-5} \cmidrule(lr){6-9}
SGCN & 0.5 & 0.37 & 130 & 51.4 $\pm$ 2.9 & 4.8 & 0.46 & 166 & 80.8 $\pm$ 1.7 \\
%\hline
SIGN & 0.7 & 0.38 & 133 & 51.6 $\pm$ 3.7 & 6.8 & 0.71 & 255 & 82.3 $\pm$ 1.8\\
%\hline
$\conftwo$ & 4.0 & 0.37 & 133 & \textbf{52.3} $\pm$ \textbf{3.5} & 224.2 & 0.72 & 476 & \textbf{83.3} $\pm$ \textbf{2.1}\\
\bottomrule
\end{tabularx}
\vspace{-10pt}
\end{table}

% \begin{table*}[tb]
% \caption{Ablation study comparing the impact of removing perturbations and the $\merge$ function (SIGN), removing perturbations and $\comb$ and $\merge$ functions (SGCN) versus $\conftwo$ on TU datasets. The results highlight the importance of graph perturbations and aggregator functions. Reported results are in terms of graph classification validation accuracy (\%). The best is in \textcolor{customgreen}{\textbf{green}}.}
% \label{tab:ablation}
% \centering
% %\scriptsize
% \renewcommand{\arraystretch}{1.1} % Adjust spacing between rows
% \setlength{\tabcolsep}{3pt}
% \begin{tabularx}{\textwidth}{@{}l *{6}{>{\centering\arraybackslash}X@{}}@{}}
% \toprule
% \textbf{Model} & \textbf{MUTAG} & \textbf{PROTEINS} & \textbf{PTC-MR} & \textbf{IMDB-B} & \textbf{IMDB-M} & \textbf{COLLAB}\\
% \midrule
% SGCN & $87.7 \pm 8.9$ & $76.2 \pm 5.3$ & $63.6 \pm 9.7$ & $74.2 \pm 3.4$ & $51.4 \pm 2.9$ & $80.8 \pm 1.77$\\
% SIGN & $87.2 \pm 9.6$ & $76.4 \pm 5.7$ & $64.5 \pm 5.8$ & $74.5 \pm 5.4$ & $51.6 \pm 3.7$ & $82.3 \pm 1.86$\\
% $\conftwo$ & \textcolor{customgreen}{$\mathbf{89.4 \pm 7.4}$} & \textcolor{customgreen}{$\mathbf{77.6 \pm 6.3}$} & \textcolor{customgreen}{$\mathbf{65.1 \pm 7.3}$} & \textcolor{customgreen}{$\mathbf{75.2 \pm 2.9}$} & \textcolor{customgreen}{$\mathbf{52.3 \pm 2.3}$} & \textcolor{customgreen}{$\mathbf{83.3 \pm 2.1}$}\\
% \bottomrule
% \end{tabularx}
% \end{table*}
However, graph perturbations in DropGNNs are not computationally efficient as for the largest dataset $\collab$, memory limitations prevented the assessment of these models on this dataset, as it requires the generation of nearly a hundred perturbations for each graph in the training phase to be fitted into the memory.\footnote{Although we encountered out-of-memory issues for DropGNN on the COLLAB dataset due to the large number of graphs and perturbations, we can avoid this issue by decreasing the batch size (e.g., from 32 to 6 in DropGCN on our tested server) to run these models on this dataset. When tested on such small batch sizes, we noticed the suboptimal performance of DropGNN. So, we reported OOM results to emphasize the computational bottleneck rather than expressiveness concerns.} In contrast, $\conftwo$ \textit{efficiently} outperforms the best-performing baseline with a 0.7\% improvement. One might wonder why $\conffour$ does not outperform all other configurations despite involving more training steps and feature extraction. We hypothesize that the main issue could be our adopted cascading approach for hyperparameter tuning, which may have led to suboptimal optimization for this configuration compared to the others.

We further compare the runtime efficiency of SE2P configurations with GCN, GIN, DropGCN, and DropGIN in Table~\ref{tab:runtimetu}. The preprocessing time of SE2P for most datasets is relatively short, ranging from a minimum of half a second for $\mutag$ to a maximum of $4$ minutes for $\collab$.\footnote{The reported preprocessing time is the time taken to preprocess the dataset and load it into memory. The process may require additional time if there is a need to write the preprocessed datasets to disk.} Across all datasets, $\confone$ and $\conftwo$ emerged as faster models for training time and total runtime (including preprocessing and training time over all epochs) than the baselines. The speedup for total runtime ranges from $3.19\times$ (in $\proteins$) to $7.90\times$ (in $\ptc$) for $\confone$, and from $2.16\times$ (in $\proteins$) to $6.02\times$ (in $\ptc$) for $\conftwo$. $\confthree$ achieves a runtime comparable to the baselines (except for $\collab$ due to a large number of graphs) while improving generalizability. $\conffour$ is the slowest model, as it involves more training time than the others. In general, if scalability of $3$-$6\times$ is desired while keeping generalizability comparable, $\conftwo$ is the best option. If one aims to maintain scalability comparable to the baselines while improving generalizability up to $1.5\%$, $\confthree$ is recommended.
% Thus, $\conftwo$ accelerates the computations up to $10\times$ faster than the other baselines.

Our SE2P methods on the OGB datasets demonstrate promising results, as shown in Table~\ref{tab:ogb}. In $\ogbmolhiv$, $\conftwo$ achieves comparable results to the baselines while offering a speedup of roughly $30\%$. $\confthree$ outperforms baseline methods but at the expense of longer training times. DropGCN, DropGIN, and $\conffour$ encountered out-of-memory issues, primarily due to the large number of graphs in this dataset ($48,127$ graphs). These issues arose from the application of message-passing over many graph perturbations (for DropGCN and DropGIN) or feature transformation over diffusion sets of all perturbations of a graph (for $\conffour$) during the training phase.
For $\ogbmoltox$, all methods utilizing node-dropout perturbations (except $\confone$, which lacks sufficient non-linearity and expressivity) outperform the two baselines without graph perturbations. For comparable performance and faster runtime, $\conftwo$ is preferred. It demonstrates roughly a $30\%$ speed improvement over the fastest baseline (GIN) and a $130\%$ speed improvement over the slowest baseline (DropGCN). If higher generalization is sought, $\confthree$ and $\conffour$ are recommended at the expense of reduced scalability.

\vskip 1mm
\noindent \textbf{Hyperparameter sensitivity analysis.}
To evaluate the sensitivity of our configurations to changes in the introduced hyperparameters, we conducted an extensive sensitivity analysis. Our results indicate that, for most datasets, SE2P configurations maintain robust performance and do not lose their comparative performance over DropGNN. Details are provided in Appendix \ref{app:sensitivity}.
\vskip 1mm
\noindent \textbf{Ablation study.} Our ablation studies intend to assess the effectiveness of perturbations and aggregator functions of $\comb$ and $\merge$ on the generalizability of $\conftwo$.\footnote{We have selected $\conftwo$ as our subject of ablation studies due to its both competitive generalizability and improved scalability/runtime, compared to baselines.} By excluding the generation of perturbations and $\merge$ from $\conftwo$, we obtain a variant SIGN \cite{sign_icml_grl2020}. For this model, we keep $\comb$ and $\pool$ the same as $\conftwo$. From SIGN, we further remove $\comb$, thus only applying one diffusion matrix (i.e., the $L$th power of adjacency matrix). The resulting model is SGCN \cite{wu2019simplifying} with the same $\pool$ function of $\conftwo$. Note that these two SIGN and SGCN variants not only allow us to conduct our ablation studies but also enable us to assess the extent of generalizability provided by the higher expressiveness of our SE2P models.  

%To compare SIGN \cite{sign_icml_grl2020} and SGCN \cite{wu2019simplifying}, which aim to speed up standard GNNs but have limitations in expressive power, with our proposed model, we conduct ablation studies on two new versions of SE2P. Firstly, we exclude the generation of perturbations in the preprocessing phase, making our model similar to SIGN. By doing this there is no need for the $\merge$ functions, and $\comb$ and $\pool$ remain as they were defined for $\conftwo$. Secondly, to mimic SGCN, we remove the generation of perturbations and consider only one diffusion matrix instead of concatenating over a set of diffusion matrices, thus removing $\comb$ and $\merge$ functions. The results of these two versions on TU datasets are shown in Table \ref{tab:ablation}.
Table \ref{tab:ablation} reports the results of our ablation studies. For all datasets, $\conftwo$ with node-dropout perturbation outperforms both SGCN and SIGN, which do not have any perturbations. This suggests that the expressive power offered by perturbations can lead to improved generalizability. %This suggests that the expressiveness offered by perturbation can easily translate to generalizability. 
Recall that the expressive power of SGCN and SIGN is bound by the $1$-WL. However, we can surpass this limitation by generating perturbations and achieving better results on benchmark datasets. The runtime of $\conftwo$ is comparable with those of  SGCN and SIGN across all datasets except for $\collab$, for which our model has a longer preprocessing time. This is due to both the large number of graphs and high average-degree graphs, which dictate a larger number of required perturbations for each graph.
\section{Conclusion and Future Work}
% We have introduced SE2P, a scalable yet expressive method for learning from graph-structured data. SE2P leverages graph perturbations and diffusion in the preprocessing stage to offer scalability while maintaining expressivity. Our framework provides flexibility in balancing scalability and expressiveness by selecting various configuration classes. Our experiments on an extensive set of benchmarks validate our approach, showing computational speedup while maintaining generalizability. Immediate future work includes designing, developing, and assessing instances of SE2P belonging to other configuration classes. Some other future directions are theoretical analysis of graph perturbations for learning through the lens of matrix perturbation theory; and adaptive and learnable selection of graph perturbation numbers and node dropout probability.
We introduced SE2P, a flexible framework with four configuration classes that balance scalability and generalizability. SE2P leverages graph perturbations and feature diffusion in the preprocessing stage and offers choices between learnable and non-learnable aggregator functions to achieve the desirable scalability-expressiveness balance. Our experiments on an extensive set of benchmarks validate the effectiveness of SE2P, demonstrating significant speed improvements and enhanced generalizability depending on the selected configuration. Future directions include exploring other graph perturbation policies beyond node dropout, providing theoretical analyses of graph perturbations through the lens of matrix perturbation theory, and developing adaptive and learnable methods for selecting the number of graph perturbations.

\bibliographystyle{ACM-Reference-Format}
\bibliography{Arxiv}
\clearpage
\appendix

\begin{table*}[t]
\caption{Number of learnable parameters.}
\label{tab:params}
\centering
\setlength{\tabcolsep}{10 pt}
\begin{tabularx}{1.0\textwidth}{@{}l@{\hskip 0.6cm}cccccccc@{}}
\toprule
\textbf{Model} & \textbf{\mutag} & \textbf{\proteins} & \textbf{\ptc} & \textbf{\imdbb} & \textbf{\imdbm} & \textbf{\collab} & \textbf{\ogbmolhiv} & \textbf{\ogbmoltox}\\
\midrule
GIN & $2440$ & $2368$ & $2656$ & $35284$ & $35614$ & $63955$ & $31054$ & $34024$\\
GCN & $3960$ & $1152$ & $4368$ & $18132$ & $18462$ & $80595$ & $13902$ & $16872$\\
DropGIN & $2440$ & $2368$ & $2656$ & $35284$ & $35614$ & N/A & N/A & $34024$\\
DropGCN & $3960$ & $1152$ & $3744$ & $18132$ & $18462$ & $\text{N/A}$ & $\text{N/A}$ & $16872$\\
\hline
$\confone$ & $464$ & $104$ & $3080$ & $21838$ & $21942$ & $1093719$ & $721$ & $930$\\
$\conftwo$ & $4914$ & $1002$ & $4210$ & $10642$ & $10659$ & $50147$ & $4033$ & $5340$\\
$\confthree$ & $9394$ & $2826$ & $6450$ & $12882$ & $17379$ & $52387$ & $8513$ & $34988$\\
$\conffour$ & $14322$ & $3594$ & $6866$ & $15186$ & $16323$ & $23139$ & $\text{N/A}$ & $41836$\\
\bottomrule
\end{tabularx}
\end{table*}
\section{Number of Learnable Parameters}
\label{app:learnable_params}
% Table~\ref{tab:params} have provided the actual number of learnable parameters for each model and dataset.
We derive equations for calculating the number of learnable parameters for each examined configuration of SE2P. We first compute the size of each learning component used in any of our configurations, then drive the size of configurations based on which components are used in them. We denote $L$ as the number of virtual layers for the diffusion step. The model size of $M$ is denoted by $|M|$ in our analyses below.

All of our models have final $MLP_f$ with $N_{f}$ hidden layers for classification; we compute its number of parameters as a function of its input size $h_0$. Assuming $C$ is the number of classes, this MLP has $\sum_{k=0}^{N_{f-1}}h_kh_{k+1}+h_{N_{f}}c$ parameters, where $h_k$ is the dimensionality of the $k^{th}$ hidden layer. For this MLP, we reduce the dimensionality of each hidden layer by a factor of $0.5$, i.e., $h_i = \frac{h_{i-1}}{2}$. Thus, $h_k = \frac{h_0}{2^k}$. In all experiments in this paper, we set $N_f=1$, so the number of parameters as a function of its input size $h_0$ is calculated by:
\begin{align}
    |MLP_f(h_0)| &= \frac{h_0}{2}(h_0+C).  
\end{align}    
Three configurations have $MLP_p$ as part of their learnable pooling functions. $MLP_p$ has $N_{p}$ hidden layers, each having fixed hidden dimensionality $h$; we compute its number of parameters as a function of its input size $h_0$. Assuming $h$ is also its output dimensionality, this MLP has 
\begin{align}
    |MLP_p(h_0)| &= h_0h + N_p h^2.
\end{align}    
DeepSet has been employed as a learning component for either $\comb$ or $\merge$ functions in two configurations $\confthree$ and $\conffour$. Assuming its inner and outer MLPs have $N_{inn}$ and $N_{out}$ hidden layers with hidden dimensionality $h$, we compute its number of parameters as a function of its input size $h_0$:
\begin{align}
    |DS(h_0)| &= h_0h + N_{inn} h^2 + (N_{out} +1) h^2\nonumber\\
    &= h_0h+ (N_{inn} + N_{out}+1)h^2.
\end{align}
From now on, we define $N^{f}_{inn}$ and $N^{f}_{out}$ as the number of hidden layers in the inner and outer MLPs of the Deepset $DS^{f}$, where $f$ specifies either the $\merge$ function or the $\comb$ function, denoted by $mer$ and $com$, respectively. 

The learnable parameters of $\confone$ are only from an MLP with one hidden layer for final classification and input size of $(L+1)d$. So the number of learnable parameters are:
\begin{align}
    |\confone| &=  |MLP_f((L+1)d)|\nonumber\\
    & = \frac{(L+1)d}{2}((L+1)d+C)  
\end{align}

$\conftwo$ has two MLPs, one for pooling $MLP_p$ with input size $(L+1)d$ and the other one for the final classification $MLP_f$ with $h$ input size. So its number of parameters is:
\begin{align}
    |\conftwo| &=  |MLP_p((L+1)d)| +  |MLP_f(h)|\nonumber\\
    & = (L+1)dh + N_p h^2 + \frac{h}{2}(h+C)  \nonumber\\
    & = (L+1)dh + (N_p+\frac{1}{2}) h^2 + \frac{Ch}{2} \nonumber\\
    & = (Ld+d+C/2)h + (N_p+\frac{1}{2}) h^2 
\end{align}

$\confthree$ has $DS^{mer}$ with input size of $(L+1)d$, $MLP_p$ with input size $h$, $MLP_f$ with input size of $h$. So its size is given by:
\begin{align}
    |\confthree| &=  |DS^{mer}((L+1)d)| +  |MLP_p(h)| +  |MLP_f(h)|\nonumber\\
        &= (L+1)dh+ (N^{mer}_{inn} + N^{mer}_{out}+1)h^2 \nonumber\\
        &\quad\quad\quad\quad\quad + (N_p+1) h^2 +  \frac{h}{2}(h+C) \nonumber\\
        &= (Ld+d+C/2)h+ (N^{mer}_{inn} + N^{mer}_{out}+ N_p +\frac{5}{2})h^2
\end{align}

$\conffour$ has $DS^{com}$ with input size $d$, $DS^{mer}$ with input size $h$, $MLP_p$ with input size $h$, and $MLP_f$ with input size of $h$. So, its number of parameters are:
\begin{align}
    |\conffour| &= |DS^{com}(d)| + |DS^{mer}(h)| + |MLP_p(h)| +  |MLP_f(h)| \nonumber\\
    & = dh + (N_{inn}^{com} + N_{out}^{com} + 1)h^2 + (N^{mer}_{inn} + N^{mer}_{out}+2)h^2 \nonumber\\
    & \quad\quad+ (N_p+1) h^2 +  \frac{h}{2}(h+C) \nonumber\\
    & = (N^{mer}_{inn} + N^{mer}_{out} + N_{inn}^{com} + N_{out}^{com} + N_p + \frac{9}{2})h^2 \nonumber\\
    &\quad\quad+ (\frac{C}2+d)h.
\end{align}
    
% \begin{align}
% \confone &= \sum_{k=0}^{N_{f}-1} \frac{(L+1)d^2}{f^{2k+1}} + \frac{(L+1)d \cdot c}{{f^{N_{f}+1}}} \\
% \conftwo &= (L+1)d \times h + (N_{p} - 1)\times h^2 + \\
% &\quad \sum_{k=0}^{N_{f}-1} \frac{h^2}{f^{2k+1}} + \frac{h \cdot c}{{f^{N_{f}+1}}} \nonumber \\
% \confthree &= (L+1)d \times h + (N^{merge}_{inner} - 1) \times h^2 + (N^{merge}_{outer} \times h^2) + \nonumber \\
% &\quad  (N_{p}) \times h^2 + \\
% &\quad  \sum_{k=0}^{N_{f}-1} \frac{h^2}{f^{2k+1}} + \frac{h \cdot c}{f^{N_{f}+1}} \nonumber \\
% \conffour &= d \times h + (N^{combine}_{inner} - 1) \times h^2 +  
%                           (N^{combine}_{outer} \times h^2) + \nonumber \\
% &\quad  (N^{merge}_{inner} \times h^2) + (N^{merge}_{outer} \times h^2) + \nonumber \\
% &\quad   (N_{p} \times h^2) + \\
% & \quad \sum_{k=0}^{N_{f}-1} \frac{h^2}{f^{2k+1}} + \frac{h \cdot c}{f^{N_{f}+1}} \nonumber
% \end{align}

\section{Hyper-parameters}
\label{app:hyperparams}
\begin{table*}[tb]
    \centering
    \caption{Selected hyperparameters for each dataset. $L$ denotes the number of virtual layers for the diffusion step. $N_{f}$ is the number of hidden layers in the final MLP. $N_{p}$ is the number of hidden layers in the $\pool$ function. $N^{mer}_{inn}$ and $N^{mer}_{out}$ are the number of hidden layers in the inner and outer MLP of the Deepset in the $\merge$ function.  $N^{com}_{inn}$ and $N^{com}_{out}$ are the number of hidden layers in the inner and outer MLP of the Deepset in the $\comb$ function. $h$ is the hidden dimensionality of all MLPs except the final MLP. The MLPs with zero hidden layers are single non-linear transformations.}
    \begin{threeparttable}
    \begin{tabularx}{\textwidth}{@{}>{\raggedright\arraybackslash}l>{\centering\arraybackslash}X>{\centering\arraybackslash}X@{}>{\centering\arraybackslash}X@{}>{\centering\arraybackslash}X@{}>{\centering\arraybackslash}X@{}>{\centering\arraybackslash}X@{}>{\centering\arraybackslash}X@{}>{\centering\arraybackslash}X@{}>{\centering\arraybackslash}X@{}>{\centering\arraybackslash}X@{}}
        \toprule
        \multirow{5}{*}{\textbf{Dataset}} & \multicolumn{10}{c}{\textbf{SE2P-C4}} \\
        \cmidrule[0.75pt]{2-11}
        & \multicolumn{8}{c}{\textbf{SE2P-C3}} & \multicolumn{2}{c}{} \\
        \cmidrule[0.75pt]{2-9}
        & \multicolumn{6}{c}{\textbf{SE2P-C2}} & \multicolumn{2}{c}{} & \multicolumn{2}{c}{}\\
        \cmidrule[0.75pt]{2-7}
        & \multicolumn{4}{c}{\textbf{SE2P-C1}} & \multicolumn{2}{c}{} & \multicolumn{2}{c}{} & \multicolumn{2}{c}{} \\
        \cmidrule[0.75pt]{2-5}
        & Batchsize & Dropout & $L$ & $N_{f}$ & $N_{p}$ & $h$ &$N^{mer}_{inn}$ & $N^{mer}_{out}$ & $N^{com}_{inn}$ & $N^{com}_{out}$\\
        \cmidrule(r){1-1}%\cmidrule(lr){2-11}
        \cmidrule(lr){2-2}\cmidrule(lr){3-3}\cmidrule(lr){4-4}\cmidrule(lr){5-5}\cmidrule(lr){6-6}\cmidrule(lr){7-7}\cmidrule(lr){8-8}\cmidrule(lr){9-9}\cmidrule(lr){10-10}\cmidrule(lr){11-11}
        $\mutag$ & $64$ & $0.5$ & $3$ & $1$ & $3$ & $32$ & $1$ & $1$ & $1$ & $2$\\
        $\proteins$ & $64$ & $0.5$ & $3$ & $1$ & $2$ & $16$ & $2$ & $2$ & $1$ & $0$\\
        $\ptc$ & $32$ & $0.0$ & $3$ & $1$ & $1$ & $32$ & $0$ & $0$ & $0$ & $0$\\
        $\imdbb$ & $32$ & $0.5$ & $2$ & $1$ & $3$ & $32$ & $0$ & $0$ & $2$ & $2$\\
        $\imdbm$ & $32$ & $0.5$ & $2$ & $1$ & $3$ & $32$ & $2$ & $2$ & $0$ & $1$\\
        $\collab$ & $32$ & $0.5$ & $2$ & $1$ & $2$ & $32$ & $0$ & $0$ & $0$ & $0$\\
        $\ogbmolhiv$ & $256$ & $0.0$ & $3$ & $1$ & $2$ & $32$ & $0$ & $1$ & $\text{N/A}$ & $\text{N/A}$\\
        $\ogbmoltox$ & $64$ & $0.0$ & $3$ & $1$ & $3$ & $64$ & $1$ & $1$ & $0$ & $0$\\
        \bottomrule
    \end{tabularx}
    \begin{tablenotes}
    \item[*] When the number of hidden layers is $0$, only a single non-linear transformation is applied.
    \end{tablenotes}
    \end{threeparttable}
    \label{tab:hyperparameters}
\end{table*}

The hyperparameters of our models (C1--C4) are shared and nested, with each more expressive configuration introducing new hyperparameters while retaining those of the less expressive model. This nested structure allows us to adopt a cascading approach for hyperparameter tuning from the $\confone$ to the $\conffour$, which helps in reducing our search space. Beginning with the least expressive configuration $\confone$, we optimized its hyperparameters and then fixed them for use in the $\conftwo$, tuning only the newly introduced hyperparameters for this configuration. We repeated this process, fixing the optimized hyperparameters of each configuration for the subsequent one ($\confthree$ and $\conffour$), and applied the hyperparameter tuning only on the new hyperparameters.
% We can proceed with this approach as in each configuration, some new hyperparameters are being introduced.
For instance, when moving from $\conftwo$ to $\confthree$, we search for two new hyperparameters: the number of inner and outer MLP layers in the DeepSet's $\merge$ function of $\confthree$, while keeping the other hyperparameters the same as in $\conftwo$. While this strategy reduces the search space, it may compromise the optimal performance of the more expressive configurations, as they might be tuned to suboptimal hyperparameters. Table \ref{tab:hyperparameters} reports our tunned hyperparameters for all four instances and datasets.

\subsection{Sensitivity Analysis}
\label{app:sensitivity}
\begin{figure*}[tb]
    \centering
    \begin{subfigure}[b]{0.46\textwidth}
        \centering
        \includegraphics[width=\textwidth]{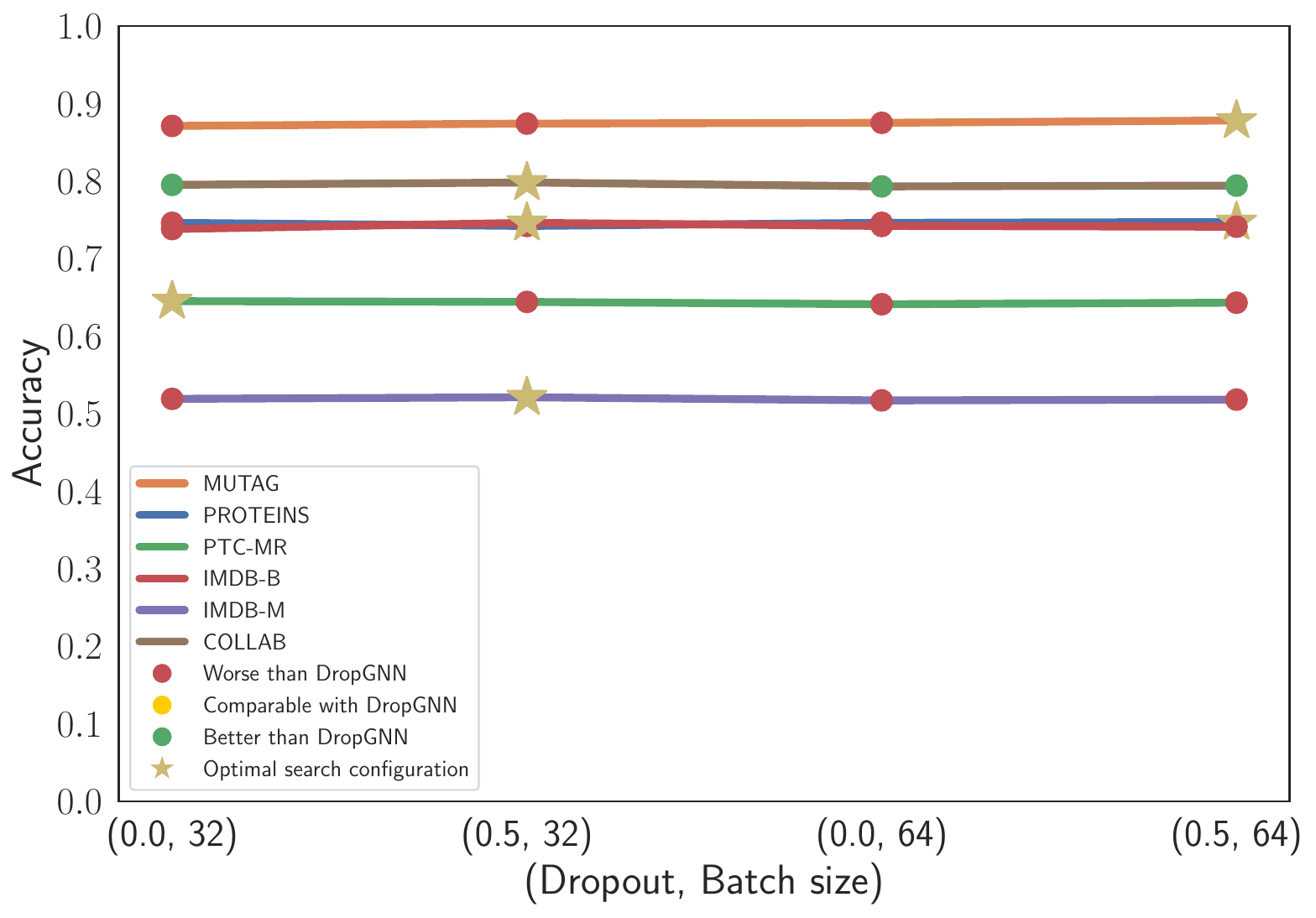}
        \caption{The hyperparameters introduced in the \confone{} configuration are dropout and batch size.\newline} %These hyperparameters are represented on the $x$-axis as (Dropout, Batch size).}
        \label{fig:sub11}
    \end{subfigure}
    \hspace{0.5cm}
    \begin{subfigure}[b]{0.46\textwidth}
        \centering
        \includegraphics[width=\textwidth]{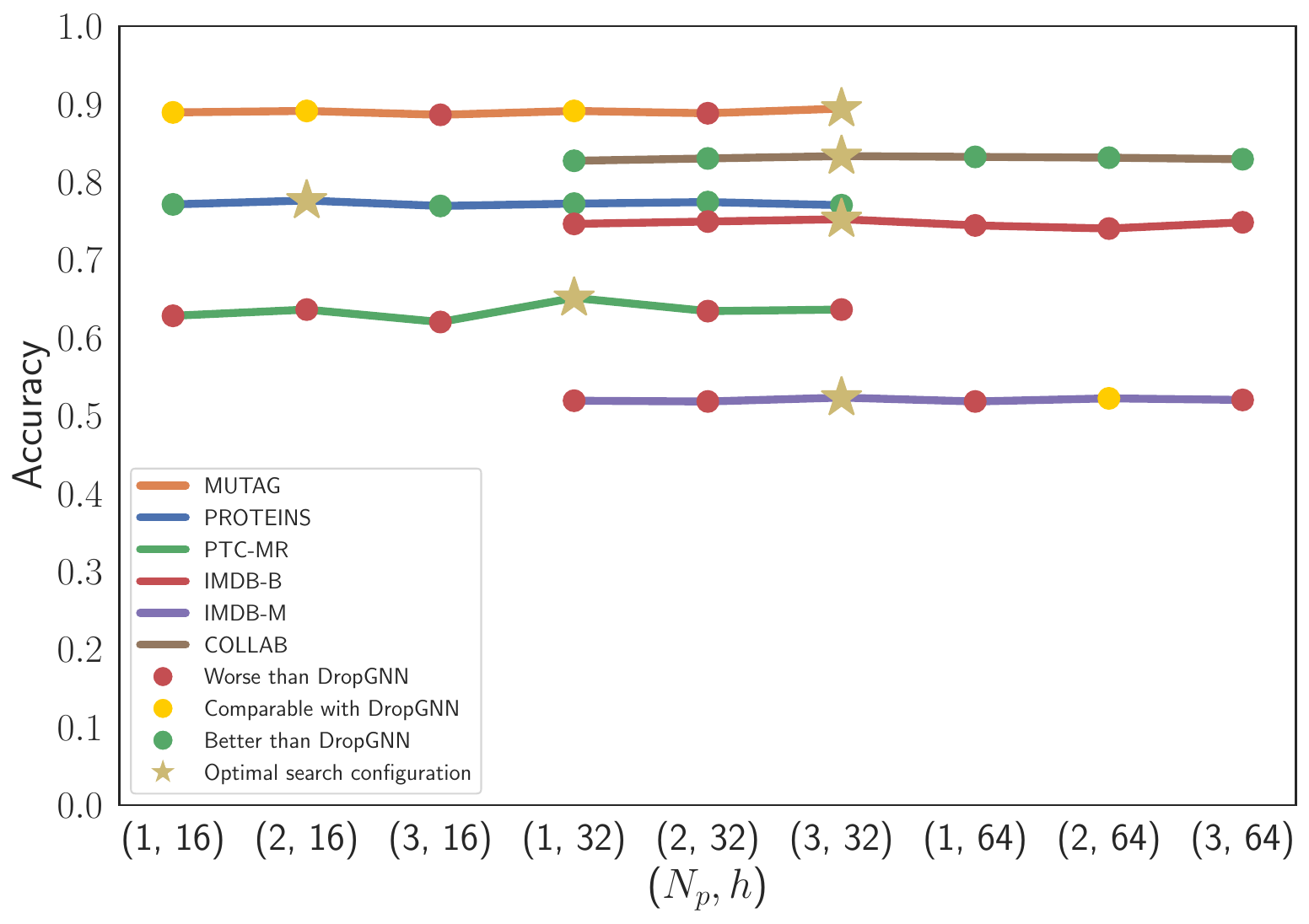}
        \caption{The hyperparameters introduced in the \conftwo{} configuration are the number of hidden layers in the $\pool$ function $N_p$ and the hidden dimensionality $h$.}% These are shown on the $x$-axis as ($N_p$, $h$).}
        \label{fig:sub22}
    \end{subfigure}

    \vspace{0.3cm}

        \begin{subfigure}[b]{0.46\textwidth}
        \centering
        \includegraphics[width=\textwidth]{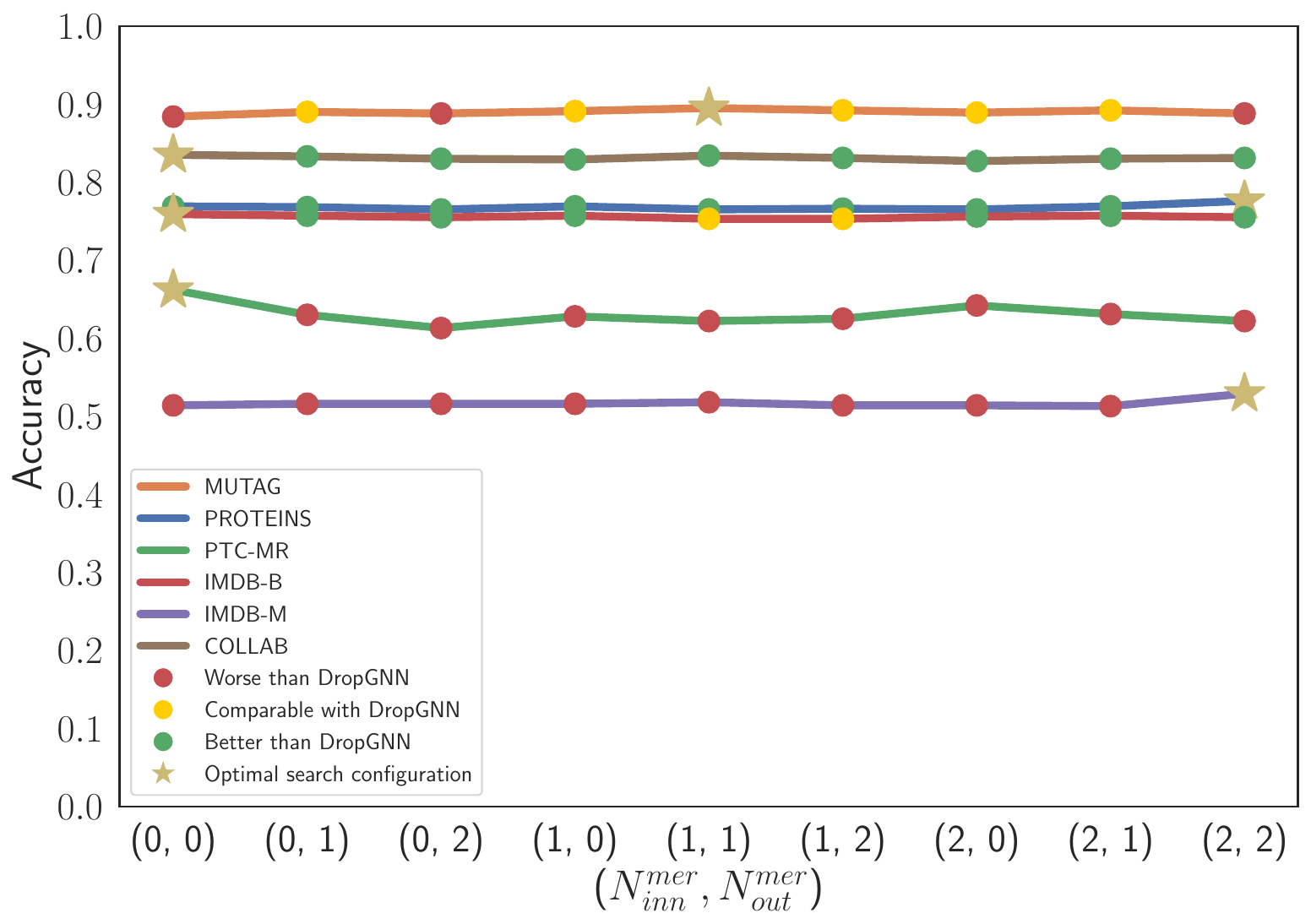}
        \caption{The hyperparameters introduced in the \conffour{} configuration are the number of hidden layers in the inner and outer MLPs of the $\merge$ function, denoted as $N^{mer}_{inn}$ and $N^{mer}_{out}$.}
        \label{fig:sub11}
    \end{subfigure}
    \hspace{1cm}
    \begin{subfigure}[b]{0.46\textwidth}
        \centering
        \includegraphics[width=\textwidth]{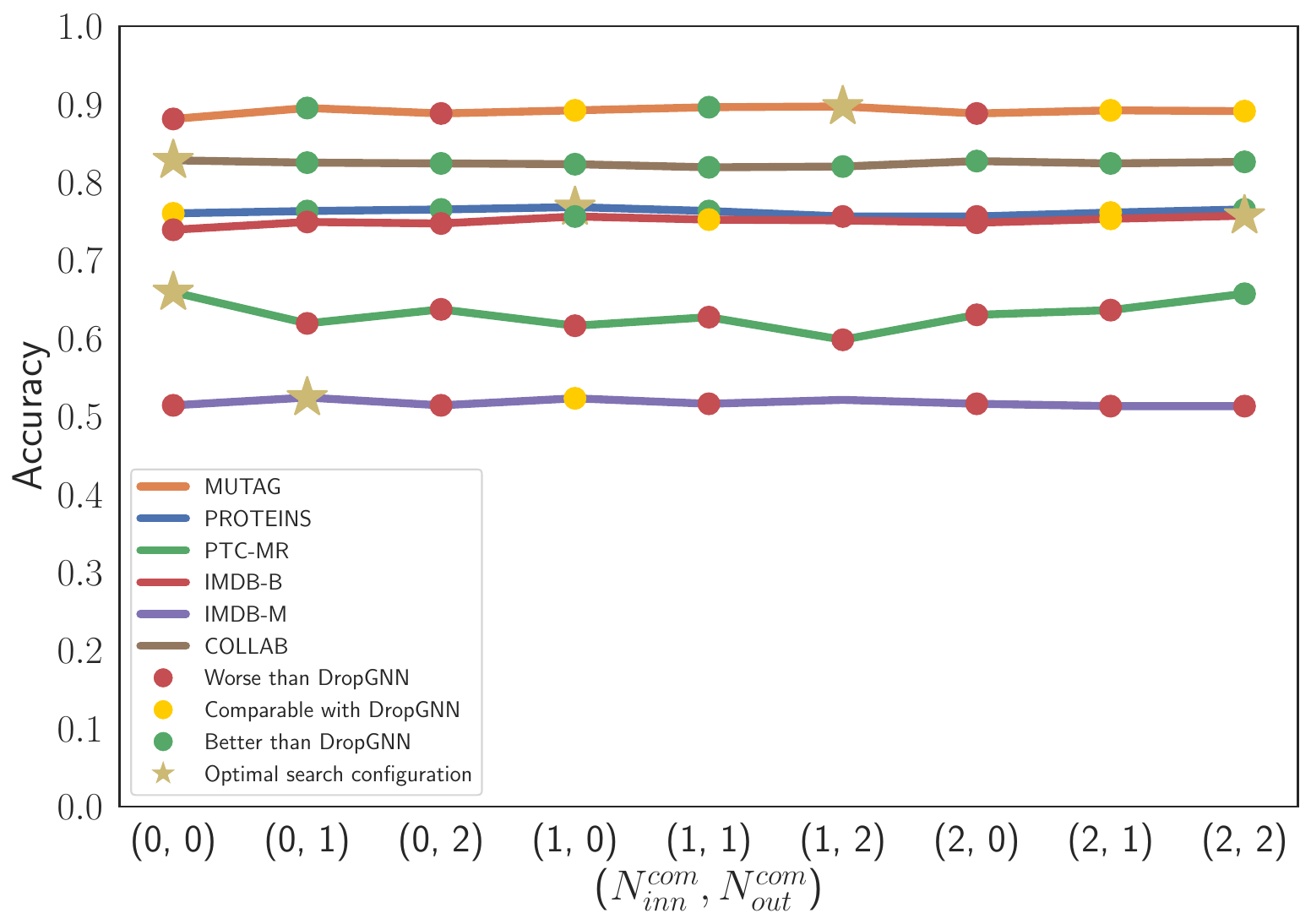}
        \caption{The hyperparameters introduced in the \conffour{} configuration are the number of hidden layers in the inner and outer MLPs of the $\comb$ function, denoted as $N^{com}_{inn}$ and $N^{com}_{out}$.}
        \label{fig:sub22}
    \end{subfigure}
    
    \caption{Sensitivity analysis on the TU datasets, investigating the impact of the introduced hyperparameters in each configuration ($x$-axis) versus the classification accuracy ($y$-axis). The star indicates the optimal hyperparameters. Each search combination is compared to DropGNN (\greencircle=better, \yellowcircle=comparable with a difference of 0.2\%, \redcircle=worse). (a) \confone{} shows insensitivity to the searched hyperparameters while usually achieving weaker results than DropGNN. (b) \conftwo{} maintains insensitivity to hyperparameters for all datasets except for the \ptc{} dataset, where sub-optimal hyperparameters show weaker results than the optimal one. (c) In \confthree{}, sub-optimal search combinations show considerably weaker performance for \imdbm{} and \ptc{}, but other datasets maintain their insensitivity to the hyperparameters, achieving comparable or better results than DropGNN. (d) \conffour{} shows relative robustness to varying the hyperparameters for all datasets, except for \ptc{}.}
    \label{fig:all}
\end{figure*}

Given the cascading approach for hyperparameter tuning, each configuration introduces two new hyperparameters. To assess the dependence of our models on hyperparameter tuning, we conduct a sensitivity analysis of the hyperparameters across all configurations for the TU datasets (see Figure \ref{fig:all}).
For \confone{}, all datasets show robust performance when varying the dropout ratio and batch size hyperparameters. Compared to DropGNN, all hyperparameter search combinations yield weaker results for all datasets except \collab{}, where DropGNN experienced OOM issues.
In \conftwo{}, we also observe insensitivity to the hyperparameters of $N_p$ and $h$ with the exception of the \ptc{} dataset, which shows some volatility when altering the introduced hyperparameters. When comparing the classification accuracy of different hyperparameter combinations with DropGNN, the combinations in \proteins{} and \collab{} consistently outperform DropGNN, while \mutag{} shows comparable results in most search combinations. 

\confthree{} introduces the number of hidden layers in the inner and outer MLP of the $\merge$ function, denoted as $N^{mer}_{inn}$ and $N^{mer}_{out}$, respectively. In this configuration, \ptc{} and \imdbm{} exhibit volatile performance when changing the hyperparameters, resulting in worse performance than DropGNN. However, in the four other datasets, most search combinations are comparable to or better than DropGNN, indicating the robust performance of \confthree{}.
The \conffour{} configuration introduces hyperparameters for the numbers of hidden layers for inner and outer MLPs (resp.) of the $\comb$ function, denoted as $N^{com}_{inn}$ and $N^{com}_{out}$ (resp.). When varying these hyperparameters, the performance of \conffour{} remains relatively stable compared to the optimal hyperparameter search, except for the \ptc{} dataset. Comparing the accuracy of different hyperparameter combinations with DropGNN shows that \conffour{} outperforms DropGNN (on all datasets, except \imdbb{}) regardless of the selected hyperparameters, implying strong robustness (or insensitivity) to hyperparameter settings.

Our sensitivity analyses show that all of our configurations for most datasets are insensitive (or robust) with regard to the hyperparameters. This means that altering hyperparameters does not result in significant performance variations or changes in their competitive performance against DropGNN. However, the \ptc{} dataset is an exception, with three configurations showing sensitivity to hyperparameters. We hypothesize that some inherent complexity in this dataset makes it challenging for our framework to learn its patterns, leading to sensitivity to hyperparameter search. Given more evidence of insensitivity to hyperparameters, we conclude that SE2P configurations are robust (or insensitive) against the hyperparameter changes.

\end{document}